\documentclass[11pt]{article}

\usepackage{comment}

\usepackage{pdfpages}

\usepackage[T1]{fontenc}
\usepackage[utf8]{inputenc}

\usepackage[top=1in, bottom=1in, left=1.15in, right=1.15in]{geometry}
\usepackage{setspace}
\usepackage{amsmath,amssymb,array}

\usepackage{authblk}

\usepackage{microtype}

\usepackage[authoryear]{natbib}
\setcitestyle{round}

\usepackage{hyperref}
\usepackage{nameref}
\hypersetup{
  unicode=true,            % allow unicode in bookmarks
  colorlinks=true,         % false = boxed links; true = colored text links
  linkcolor=black,          % color for internal links (sections, toc)
  citecolor=blue,          % color for citations
  urlcolor=blue,           % color for URLs
  pdfborder={0 0 0}        % do not draw a border (extra safety)
}

\usepackage{graphicx}
\usepackage{subcaption}
\usepackage{float}
\usepackage{booktabs}
\usepackage[font=small,labelfont=bf]{caption}

\title{Classifier Calibration at Scale: An Empirical Study of Model-Agnostic Post-Hoc Methods}
\author[1]{Valery Manokhin\thanks{\texttt{Valery.Manokhin.2015@live.rhul.ac.uk}}}
\author[2]{Daniel Grønhaug\thanks{\texttt{dfgronha@math.uio.no}}}
\affil[1]{Independent Researcher}
\affil[2]{University of Oslo}
\date{} % remove date

\begin{document}

\maketitle
\thispagestyle{empty}
\begin{abstract}
We study model-agnostic post-hoc calibration methods intended to improve probabilistic predictions in supervised binary classification on real i.i.d.\ tabular data, with particular emphasis on conformal and Venn-based approaches that provide distribution-free validity guarantees under exchangeability. We benchmark 21 widely used classifiers, including linear models, SVMs, tree ensembles (CatBoost, XGBoost, LightGBM), and modern tabular neural and foundation models, on binary tasks from the TabArena-v0.1 suite using randomized, stratified five-fold cross-validation with a held-out test fold. Five calibrators---Isotonic regression, Platt scaling, Beta calibration, Venn--Abers predictors, and Pearsonify---are trained on a separate calibration split and applied to test predictions. Calibration is evaluated using proper scoring rules (log-loss and Brier score) and diagnostic measures (Spiegelhalter's \(Z\), ECE, and ECI), alongside discrimination (AUC-ROC) and standard classification metrics.

Across tasks and architectures, Venn--Abers predictors achieve the largest average reductions in log-loss, followed closely by Beta calibration, while Platt scaling exhibits weaker and less consistent effects. Beta calibration improves log-loss most frequently across tasks, whereas Venn--Abers displays fewer instances of extreme degradation and slightly more instances of extreme improvement. Importantly, we find that commonly used calibration procedures, most notably Platt scaling and isotonic regression, can systematically degrade proper scoring performance for strong modern tabular models. Overall classification performance is often preserved, but calibration effects vary substantially across datasets and architectures, and no method dominates uniformly. In expectation, all methods except Pearsonify slightly increase accuracy, but the effect is marginal, with the largest expected gain about \(0.008\%\). Code and artifacts are available at \href{https://github.com/valeman/classifier_calibration/tree/release-v1.0}{github.com/valeman/classifier\_calibration}.
\end{abstract}

\newpage

\tableofcontents

\newpage
\section{Introduction}
This paper studies model-agnostic post-hoc calibration methods for producing better-calibrated class probability vectors in supervised binary classification on real i.i.d.\ tabular data.
We focus on methods applicable to popular machine learning models, including classical models (e.g., support vector machines), decision tree models (e.g., XGBoost, CatBoost), and neural networks (e.g., ModernNCA).
Our key objectives include: 
\begin{itemize}
    \item \textbf{Identify Robust Calibration Techniques}: \newline
    Identify methods that consistently yield well-calibrated outputs, serving as strong starting points for practitioners.
    \item \textbf{Model-Agnostic Approaches}: \newline 
    All methods explored are general approaches applicable across a variety of models and datasets. We do not explore methods specific to any architecture. 
    \item \textbf{Preserve Overall Performance}: \newline 
    Calibration is not considered in isolation, we aim to preserve overall performance.
\end{itemize}

Calibration is the degree of which class scores reflect actual probabilities. If the score of class $j$ is 0.8, you should expect class $j$ to be the true class 80\% of the time \citep{pavlovic_understanding_2025}.  Calibration is critical for reliable decision making in risk-sensitive domains (healthcare, finance, etc.) \citep{chen_calibration_2018,fonseca_calibration_2017} and for tasks like model ensembling \citep{zhang_mix-n-match_2020} , cost-sensitive learning \citep{yang_cost-aware_2025}, and uncertainty quantification \citep{shen_post-hoc_2023}. However, calibration is essential not only in risk-sensitive domains but also in various machine learning applications, including ranking systems \citep{kweon_obtaining_2022} and online advertising \citep{wei_posterior_2024}, where accurate probabilistic predictions are crucial for optimal performance.

To identify which calibration techniques consistently deliver well-calibrated outputs, we employ an empirical research design. We evaluate twenty-one popular classifiers, namely Class prior, Support vector machine, Logistic Regression, Linear Discriminant Analysis, Naïve bayes, K-Nearest Neighbours, RandomForest, Gradient Boosting Classifier, Histogram Gradient Boosting, ExtraTrees, Explainable Boosting Machine, Catboost, XGBoost, LightGBM, ModernNCA, TabTransformer, TabICL, TabPFN v2, TabM, Multilayer Perceptron and Real Multilayer Perceptron. We evaluate all methods on binary tasks from the TabArena-v0.1 dataset suite using randomized, stratified five-fold cross-validation, where four folds are used for training and one fold is held out for out-of-sample testing. We then apply five post‑hoc calibrators generated by holding out a calibration set from the training folds. The calibrations are Isotonic regression, Platt scaling, Beta calibration, Venn‑Abers predictors, and Pearsonify. Thereafter, we measure changes in calibration on the held-out test fold after applying the calibrators using Brier score, Spiegelhalter Z-statistic, Log-loss, ECE, and ECI. Afterwards we control for changes in discriminatory ability (AUC‑ROC), classification (accuracy, recall, precision and F1 score), and computational cost (training/inference time, CPU/RAM usage). Methods that consistently improve calibration without degrading overall performance are highlighted. Not all measurements and models are presented in our results, to view specific measures on specific models examine the artifacts in our archive. These are available on our \href{https://github.com/valeman/classifier_calibration/tree/release-v1.0}{GitHub repository}.

Among the post-hoc methods studied—Isotonic regression, Platt scaling, Beta calibration, Venn–Abers predictors, and Pearsonify—we observe substantial variation in their effects across models and datasets. In particular, Platt scaling and isotonic regression can degrade calibration and proper scoring performance for strong modern tabular models. By contrast, Beta calibration and Venn–Abers predictors more consistently improve calibration relative to out-of-the-box baselines, although no method dominates uniformly and overall classification performance is typically preserved.

\section{Results}

The experiment described in section \ref{sec:experiment} produce logs which we leverage to generate the statistics we'll present here. These logs can be found in the \href{https://github.com/valeman/classifier_calibration/tree/release-v1.0/src/archive}{archive}. We'll begin by exploring the performance of the base models across all datasets and folds. Then we'll examine how each calibration method affect this performance. Lastly, we'll comment on inter-model variance.

\begin{figure}[H]
    \centering
    \includegraphics[width=0.7\linewidth]{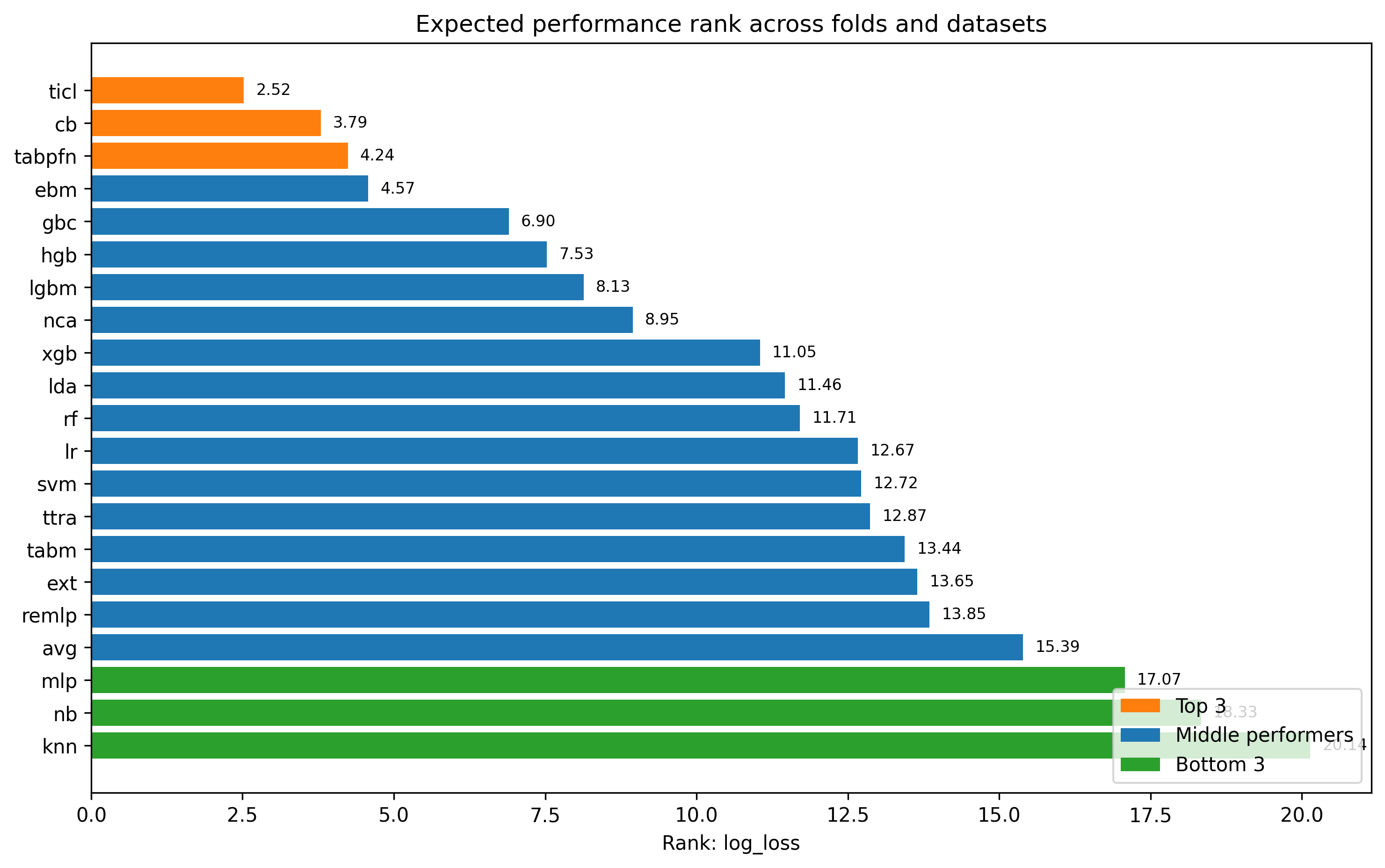}
    \caption{The expected ranking of each model by log-loss, across datasets and folds. No calibration. Evaluated on an out-of-sample test set. The best rank is 1 and the worst is 21.}
\end{figure}
The foundation models and tree based models tend to outperform with regards to log-loss. MLP, NB and KNN are the bottom performers and tend to do worse then AVG. TICL leads with an expected rank of 2.52, which is 1.27 lower then the runner up CB. CB, TABPFN and EBM are within a whole rank of each-other. GBC and the rest follow up with more the 2.33 higher expected rank then EBM. 

\begin{figure}[H]
    \centering
    \includegraphics[width=0.7\linewidth]{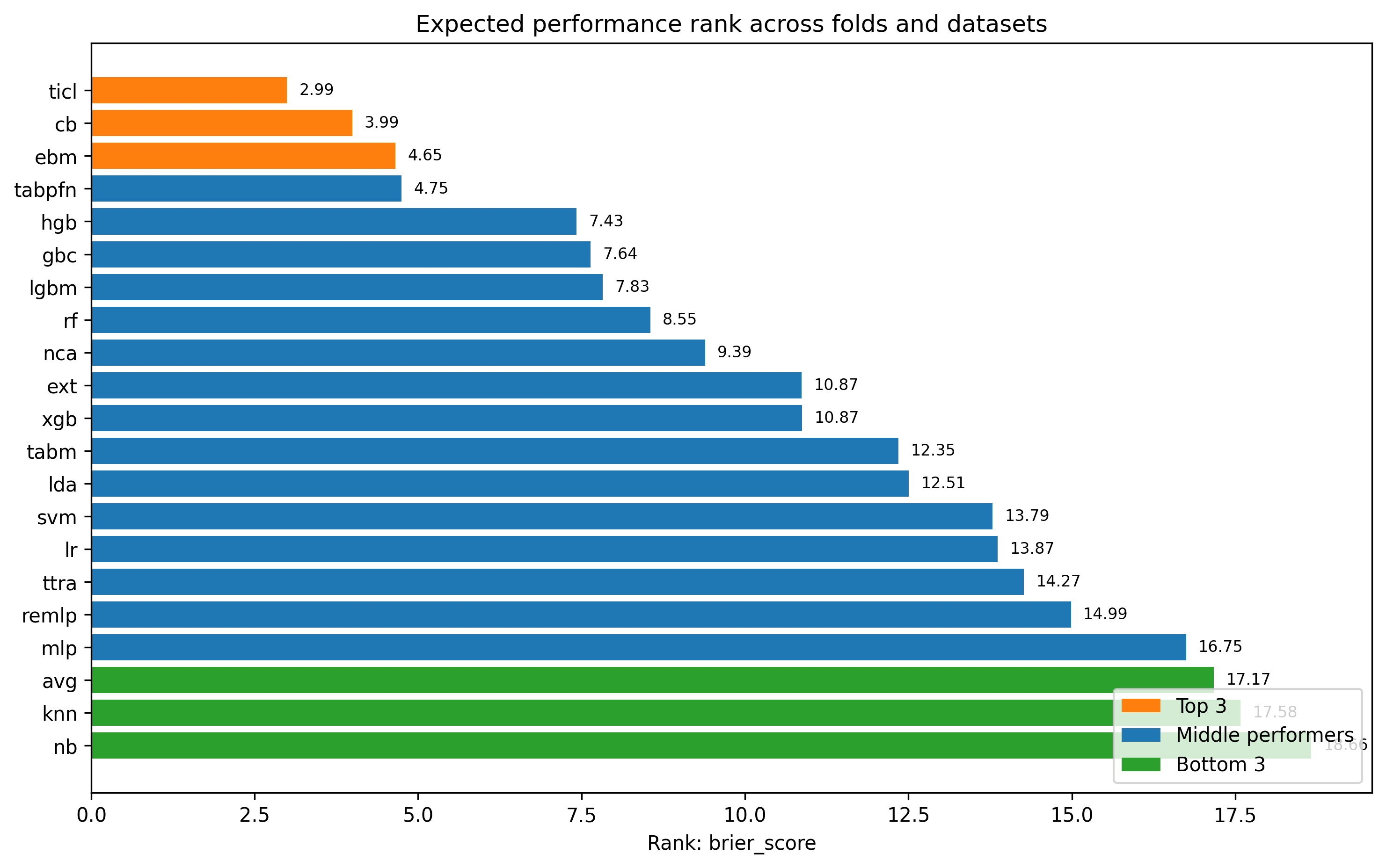}
    \caption{The expected ranking of each model by brier score, across datasets and folds. No calibration. Evaluated on an out-of-sample test set. The best rank is 1 and the worst is 21.}
\end{figure}
As for brier score, the picture is similar too log-loss. But values vary. TICL leads with an expected rank of 2.99, CB second with 3.99. CB, EBM and TABPFN are within a whole rank of each-other before the rest follow after a gap of 2.68 in expected rank.

\begin{figure}[H]
    \centering
    \includegraphics[width=0.7\linewidth]{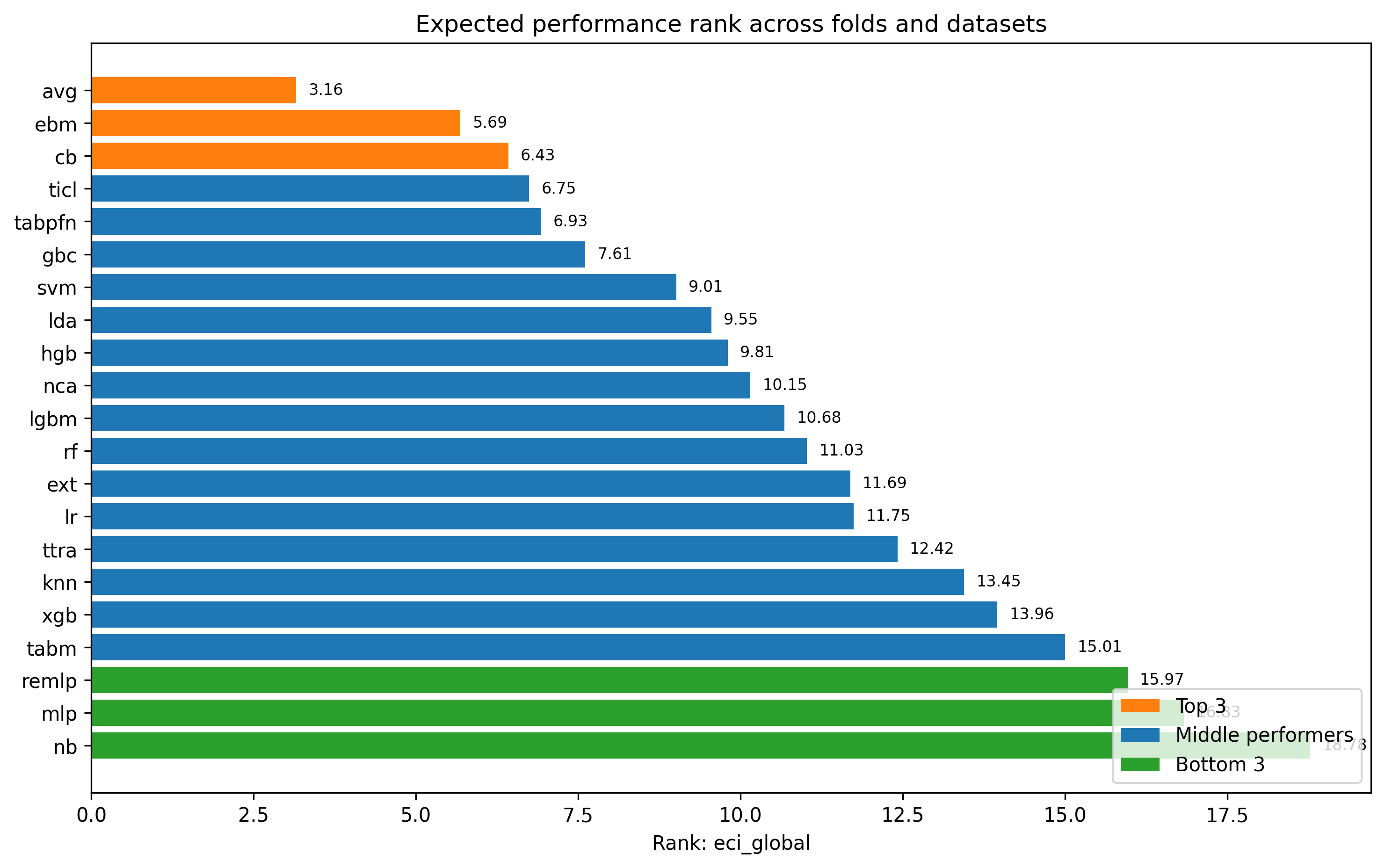}
    \caption{The expected ranking of each model by ECI global, across datasets and folds. No calibration. Evaluated on an out-of-sample test set. The best rank is 1 and the worst is 21.}
\end{figure}
The highest performer is AVG with an expected rank of 3.16. This illustrates the aforementioned issue with ECE based measures. Predicting the constant expectation or the empirical class distribution minimizes the measure. Notably, EBM, CB, TICL and TABPFN still make up the top ranks. However, now GBC is a closer contender.

\begin{figure}[H]
    \centering
    \includegraphics[width=0.7\linewidth]{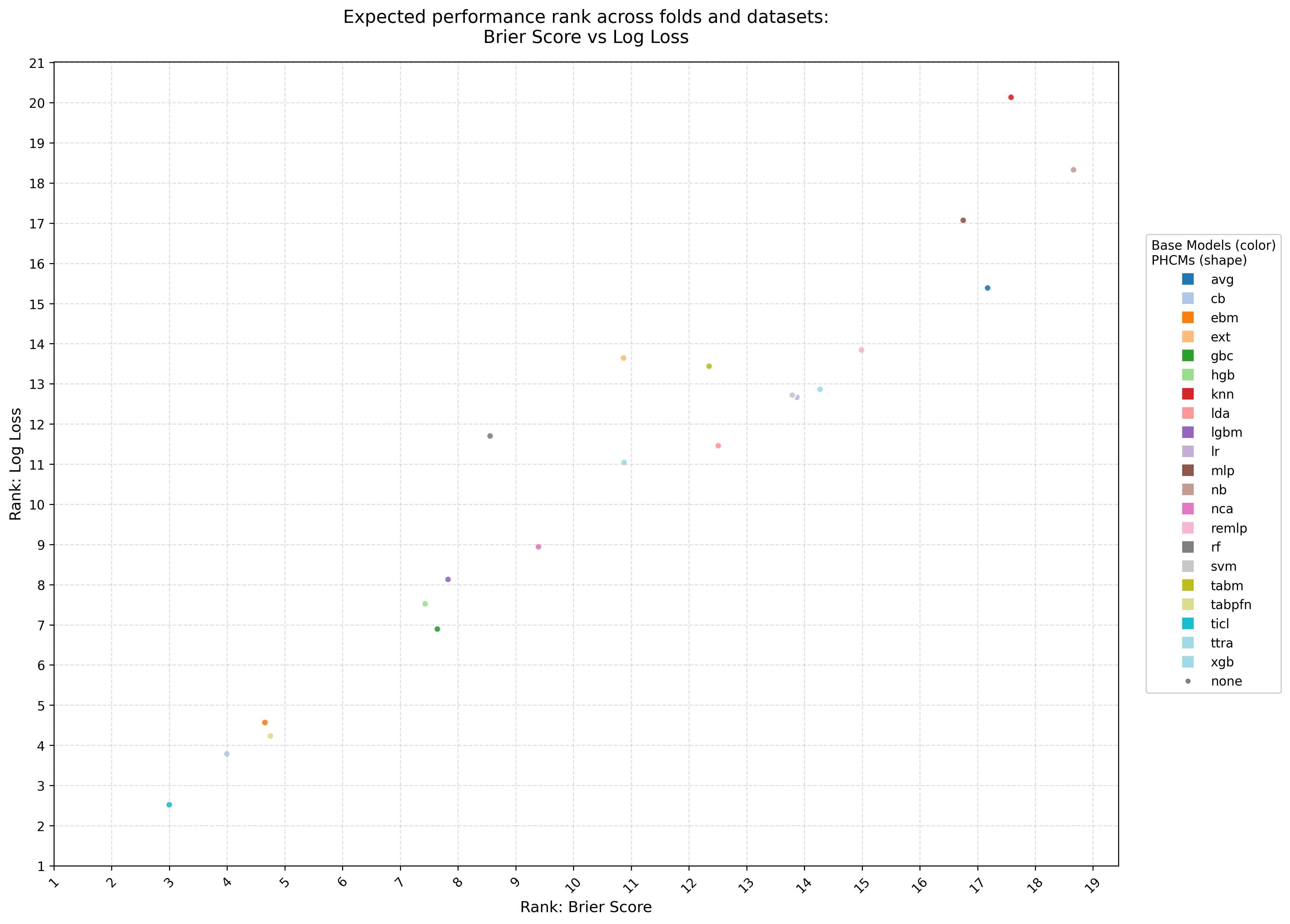}
    \caption{The expected ranking of each model by brier score and log-loss, across datasets and folds. No calibration. Evaluated on an out-of-sample test set. The best rank is 1 and the worst is 21.}
\end{figure}
A model's log-loss and brier score rank are relatively linear with some outliers. Notably log-loss penalizes large deviations between the true class and the assigned probability more then the brier score. A lower brier score rank relative to log-loss rank could therefore be interpreted as having more predictions closer to the true class, but also a higher prevalence of large deviations. KNN for example ranks higher in log-loss then in brier score, which we interpret as a larger prevalence of egregious mistakes (i.e. assigning a probability of $0$ to the true class.).

\begin{figure}[H]
    \centering
    \includegraphics[width=0.7\linewidth]{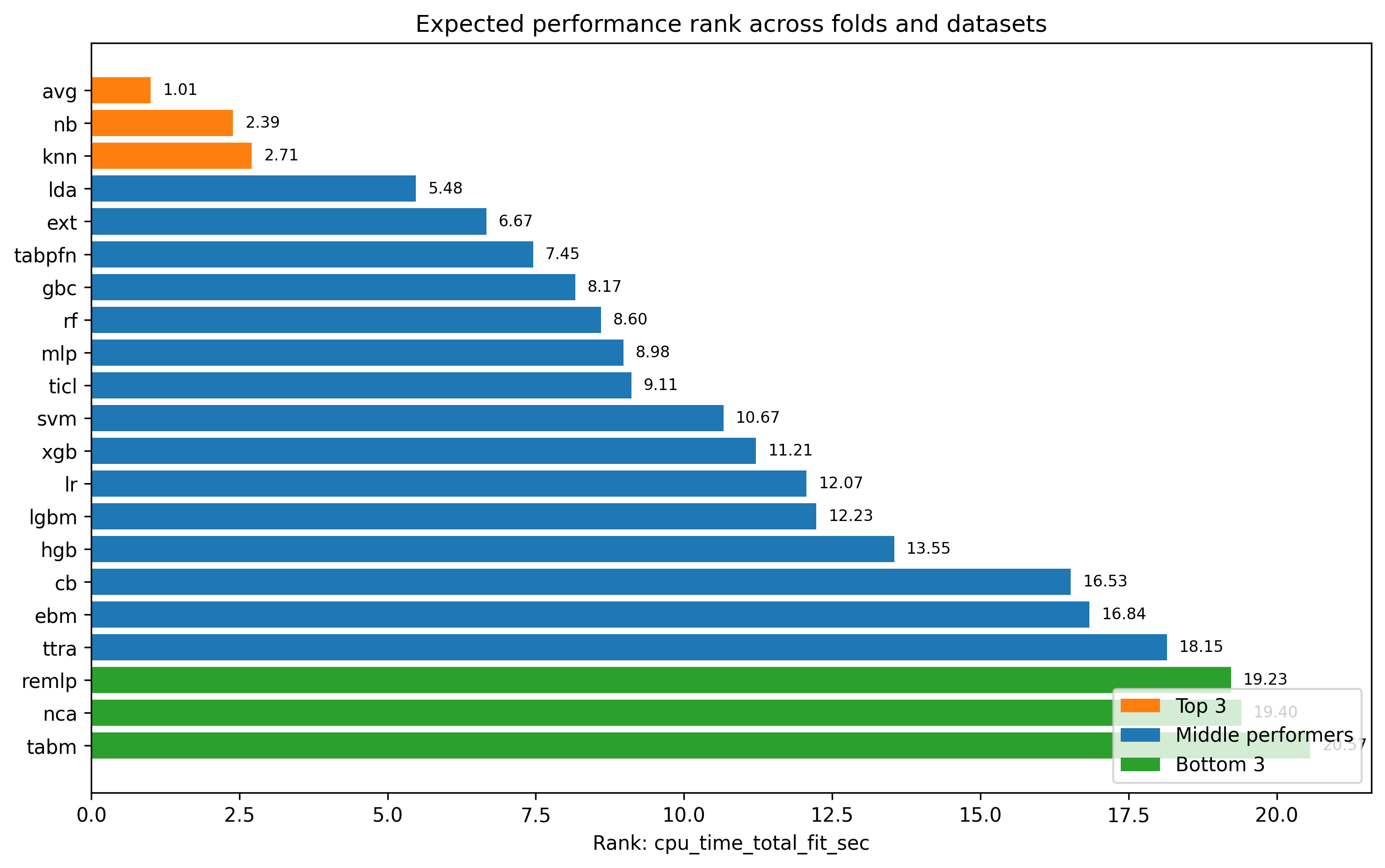}
    \caption{The expected ranking of each model by total CPU seconds during training, across datasets and folds. No calibration. Evaluated during training. The best rank is 1 and the worst is 21.}
\end{figure}
AVG ranks 1st with an expected rank of 1.01 and is the model which consumes the least compute during training. Followed by NB, KNN and LDA with respective expected ranks of 2.39, 2.71 and 5.48. The bottom three are TABM, NCA and REMLP. TABPFN ranks 6th with an expected rank of 7.45 \footnote{Note, for TABPFN we've used settings which shift compute to inference since the docs advised as such when few instances of inference are expected.}. CB and EBM rank 16th and 17th respectively with an expected rank of 16.53 and 16.84. TICL ranks 10th with an expected rank of 9.11.

\begin{figure}[H]
    \centering
    \includegraphics[width=0.7\linewidth]{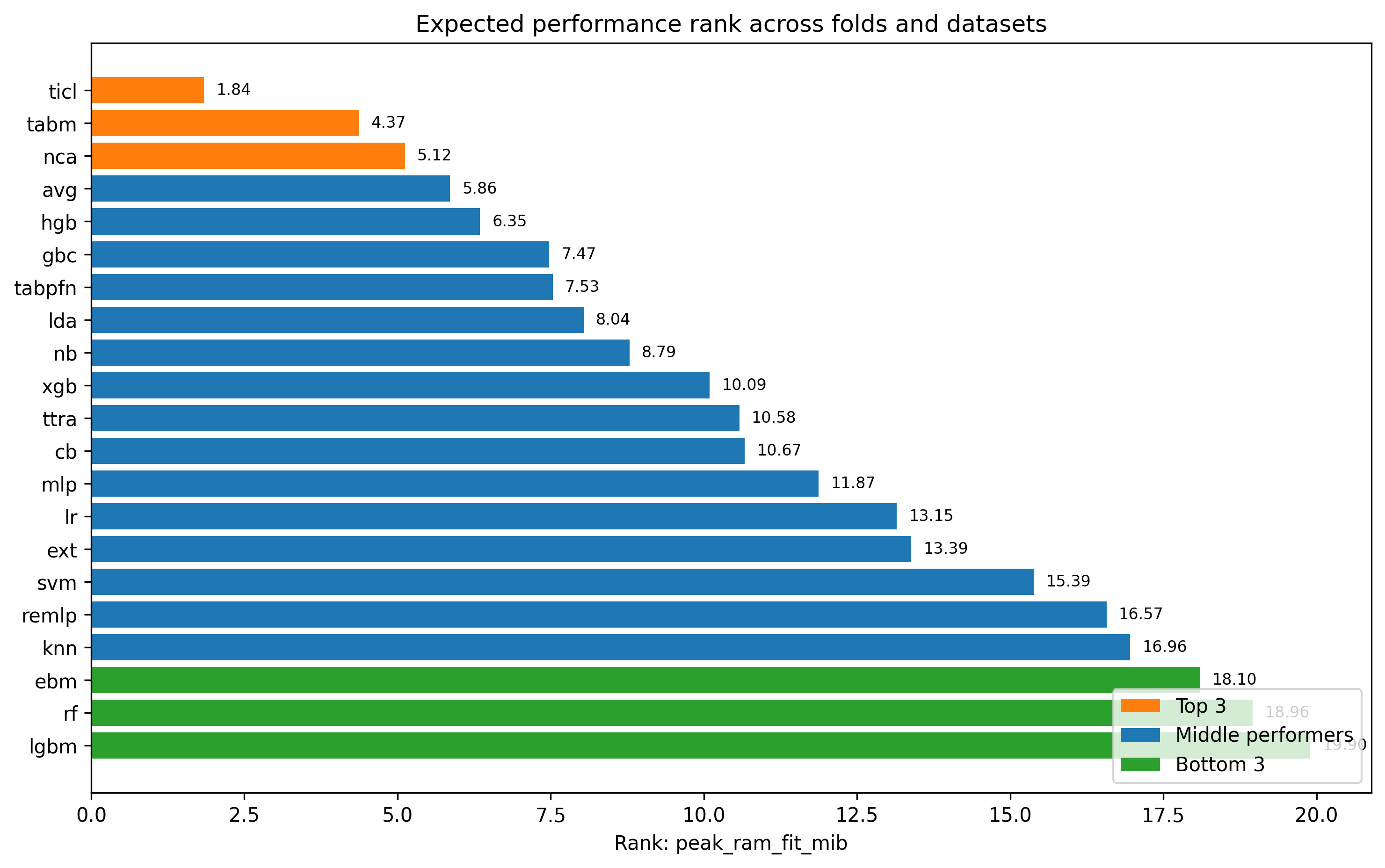}
    \caption{The expected ranking of each model by peak RAM usage during training, across datasets and folds. No calibration. Evaluated during training. The best rank is 1 and the worst is 21.}
\end{figure}
In terms of peak RAM usage during training TICL leads with an expected rank of 1.84, while TABM, NCA and AVG follow with 4.37, 5.12 and 5.86. The bottom performers are KNN, EBM, RF and LGBM. While TABPFN has an expected rank of 7.53 and CB has 10.67.

\begin{figure}[H]
    \centering
    \includegraphics[width=0.7\linewidth]{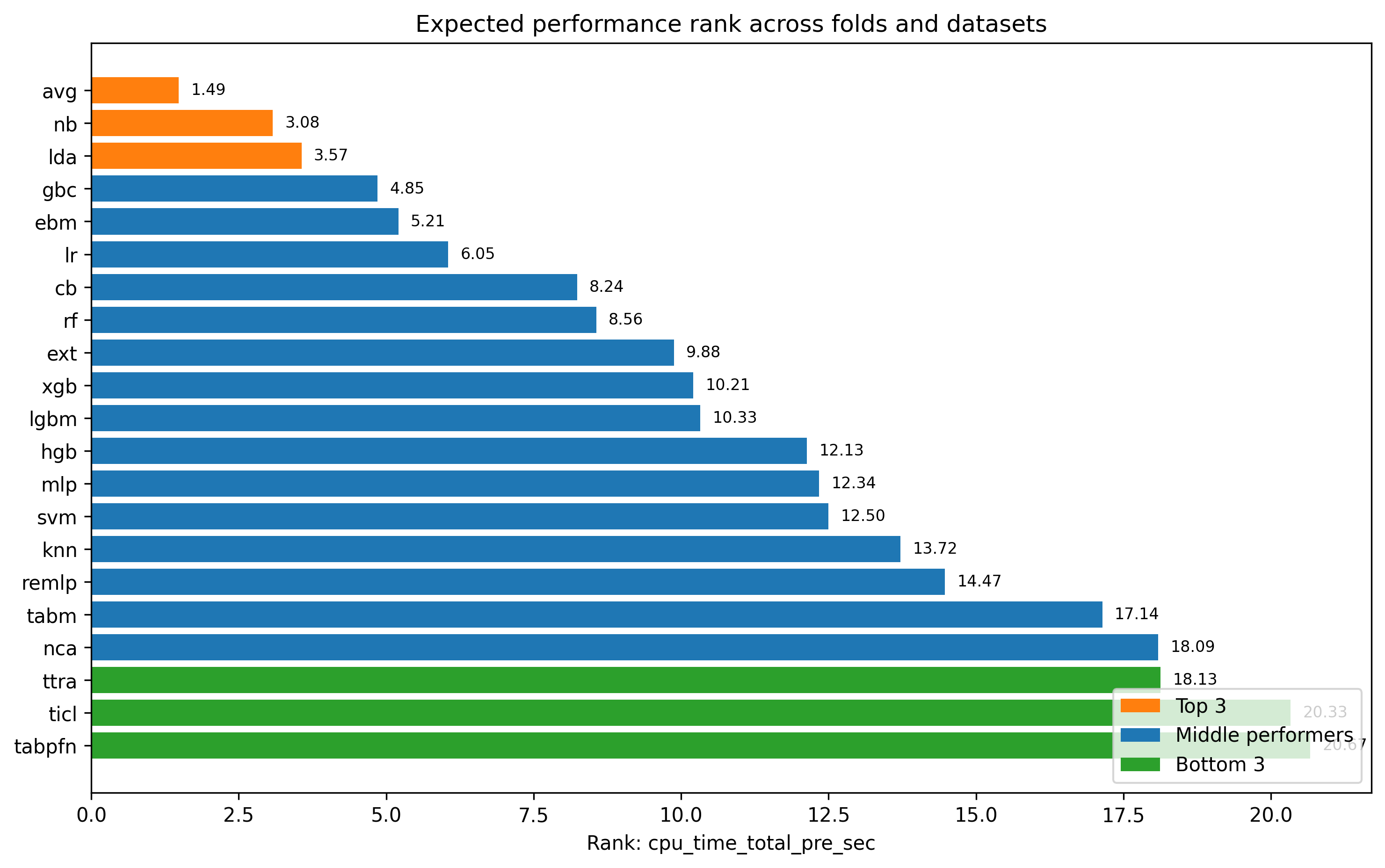}
    \caption{The expected ranking of each model by total CPU seconds during inference, across datasets and folds. No calibration. Evaluated during inference. The best rank is 1 and the worst is 21.}
\end{figure}
Notably TICL and TABPFN almost always rank last with regards to CPU seconds during inference with respective expected ranks of 20.33 and 20.67 out of 21 models. Top three are AVG, NB and LDA with respective 1.49, 3.08 and 3.57 as expected ranking. EBM and CB rank 5th and 7th with expected rankings of 5.21 and 8.24.  

\begin{figure}[H]
    \centering
    \includegraphics[width=0.7\linewidth]{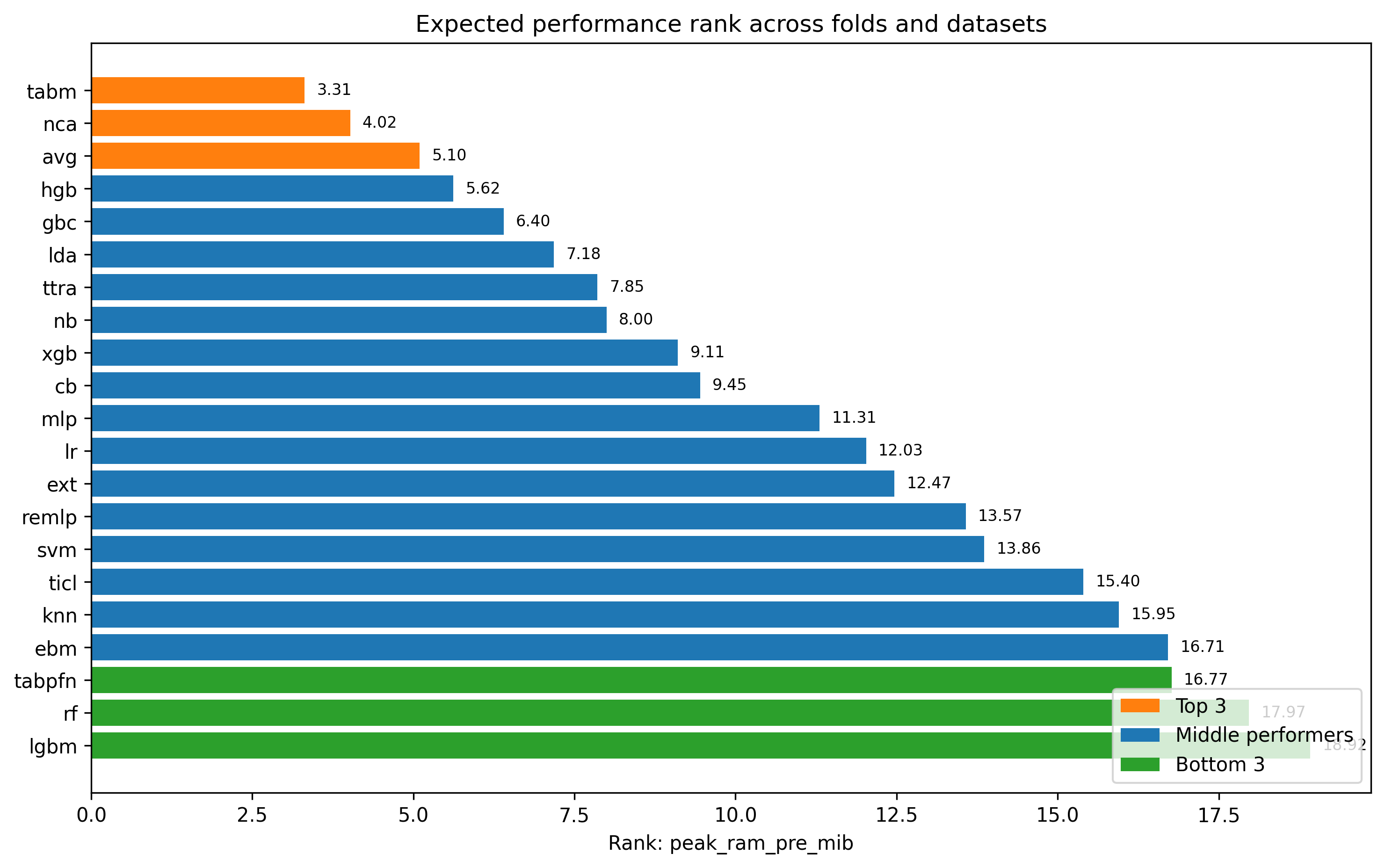}
    \caption{The expected ranking of each model by peak RAM usage during inference, across datasets and folds. No calibration. Evaluated during inference. The best rank is 1 and the worst is 21.}
\end{figure}
In terms of peak RAM usage during inference TABM, NCA and AVG lead with 3.31, 4.02 and 5.10 in expected rank. TABPFN, RF and LGBM come last. TICL has an expected ranking of 15.4, CB has 9.45 and EBM has 16.71.

\begin{figure}[H]
    \centering
    \includegraphics[width=0.7\linewidth]{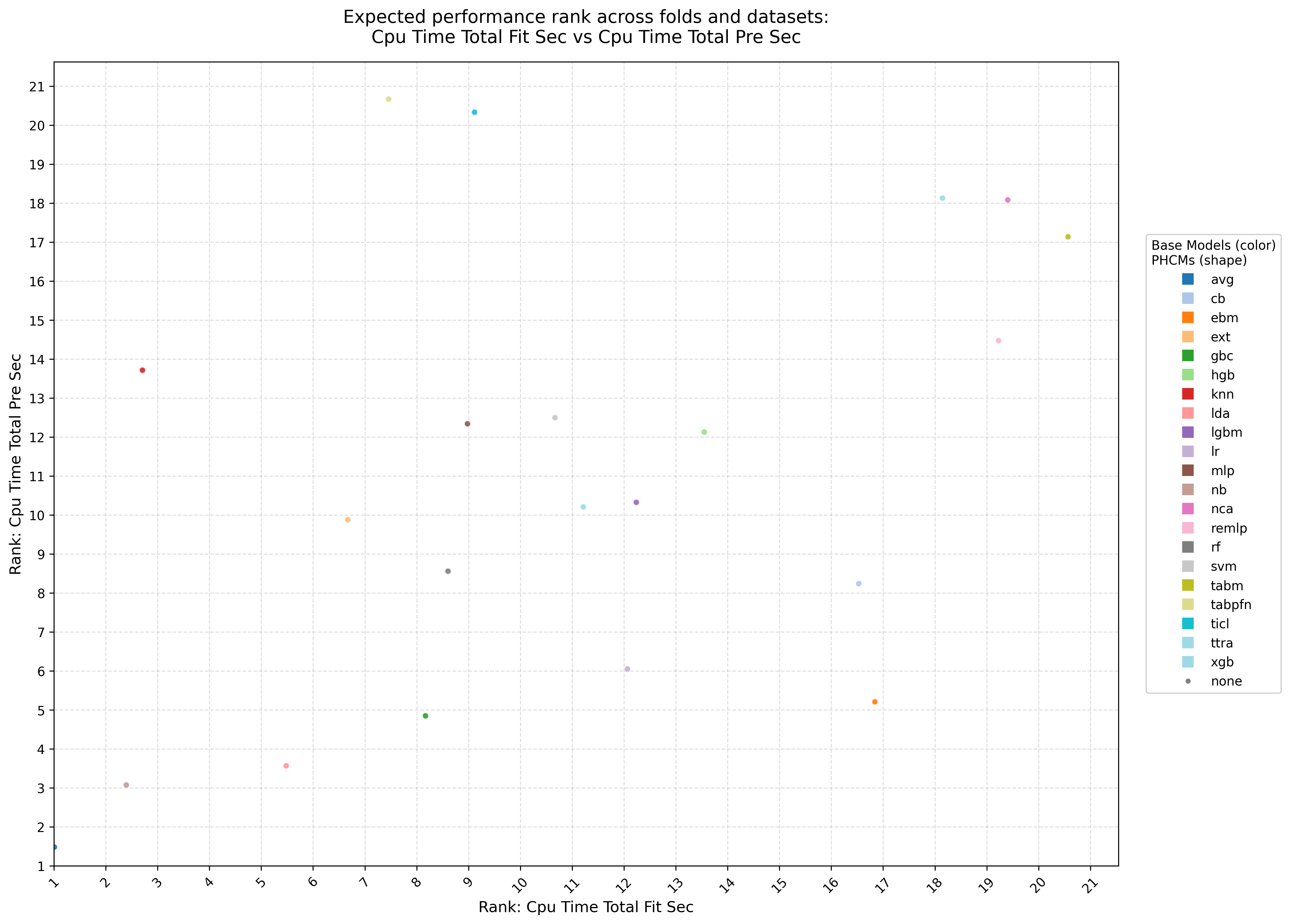}
    \caption{The expected ranking of each model by total CPU seconds during training and inference, across datasets and folds. No calibration. The best rank is 1 and the worst is 21.}
\end{figure}
The lowest compute consumption during training and inference is given by AVG, followed by NB, LDA, GBC and RF. The worst are given by NCA, TABM and TTRA. TICL ranks in the middle wrt. training but very low wrt. inference. TABPFN lies close to TICL but with somewhat worse inference rank and better training rank \footnote{Remember pre-trained models have consumed compute resources not reported here.}. CB ranks lower-mid wrt. to training but upper-middle wrt. inference. Lastly EBM provides lower-mid training cost but with good inference cost. 

\begin{figure}[H]
    \centering
    \includegraphics[width=0.7\linewidth]{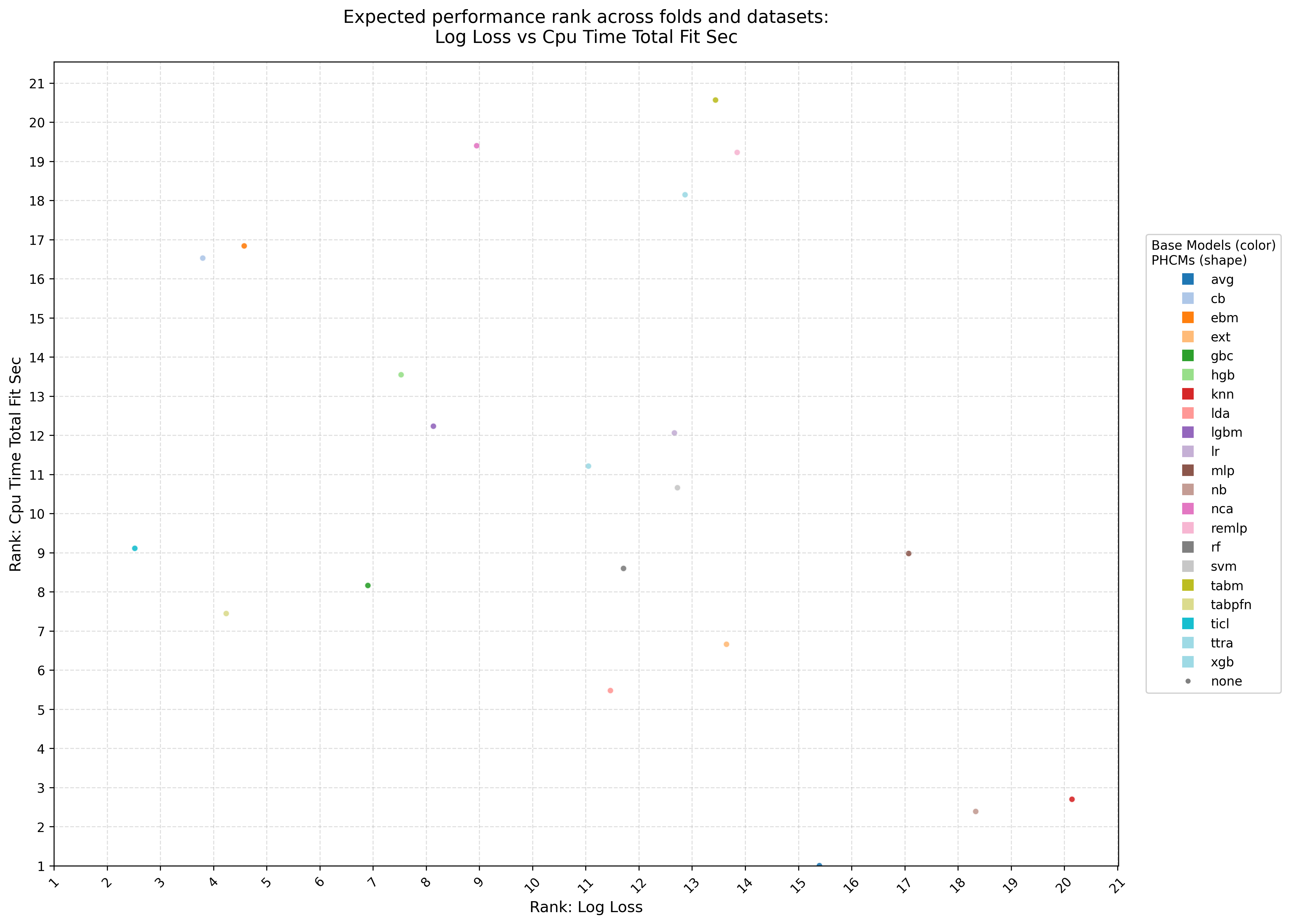}
    \caption{The expected ranking of each model by total CPU seconds during training and log-loss, across datasets and folds. No calibration. The best rank is 1 and the worst is 21.}
\end{figure}
TICL gives best log-loss but at a somewhat higher cost to training compute. in contrast, TABPFN provides somewhat worse log-loss but you save training compute. CB and EBM provide good log-loss but are on the worse end wrt. to training compute. 

\begin{figure}[H]
    \centering
    \includegraphics[width=0.7\linewidth]{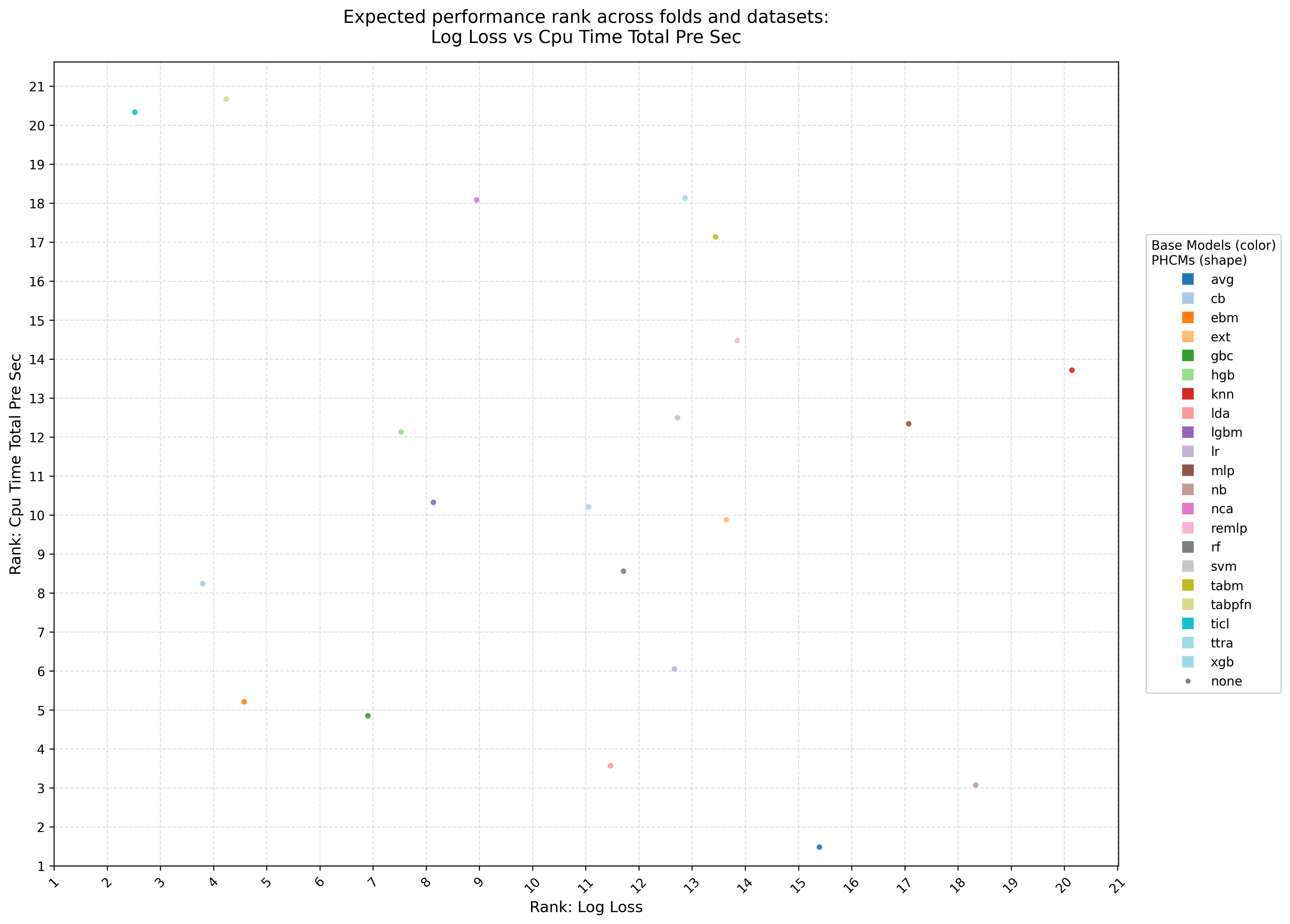}
    \caption{The expected ranking of each model by total CPU seconds during inference and log-loss, across datasets and folds. No calibration. The best rank is 1 and the worst is 21.}
\end{figure}
The best balance between log-loss and inference compute is given by EBM, GBC and CB. While, TABPFN and TICL provide better log-loss but at bottom rank inference compute costs.

\begin{figure}[H]
    \centering
    \begin{subfigure}{0.48\linewidth}
        \centering
        \includegraphics[width=\linewidth]{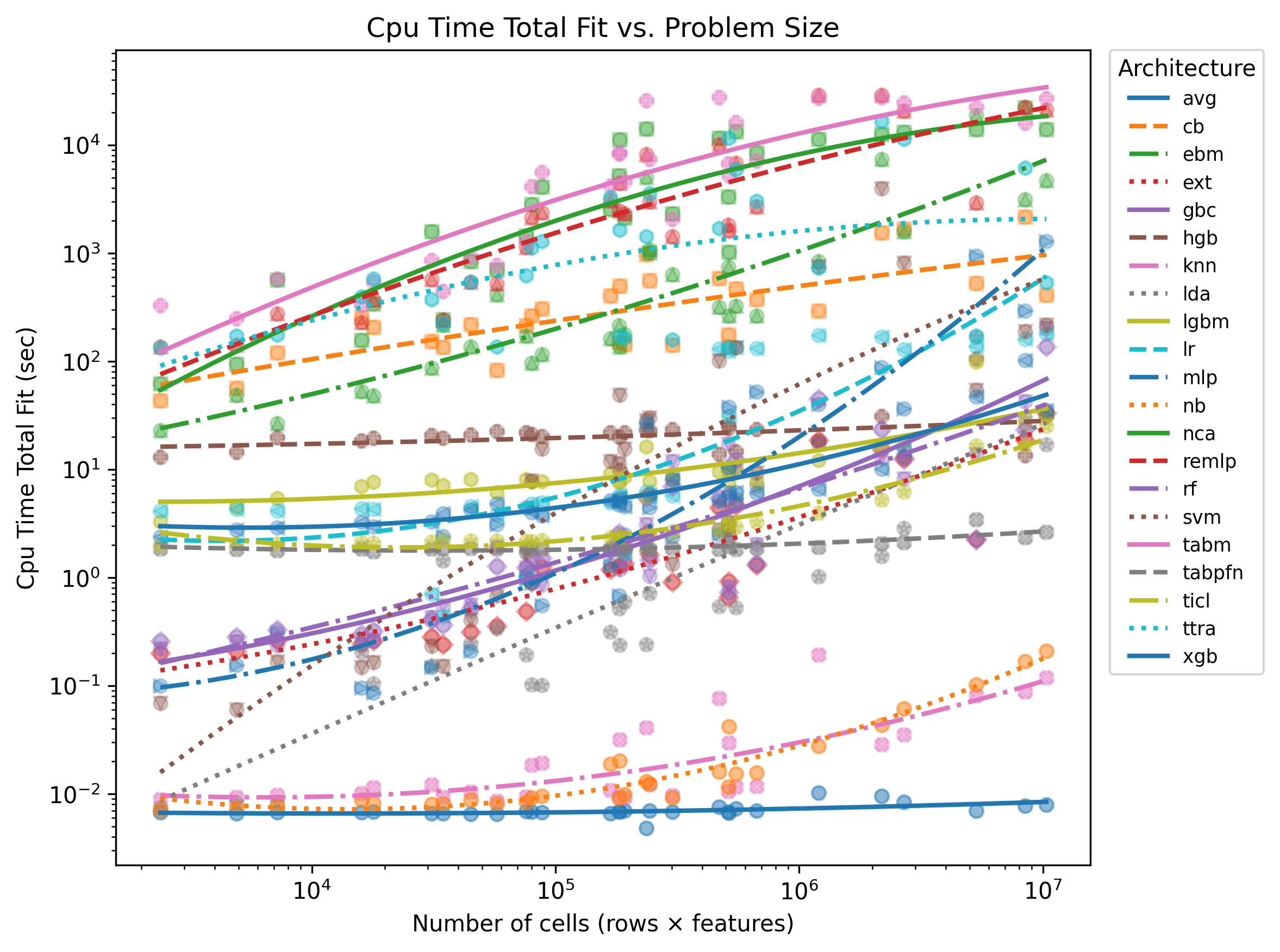}
        \caption{Training CPU time vs. cells}
    \end{subfigure}
    \hfill
    \begin{subfigure}{0.48\linewidth}
        \centering
        \includegraphics[width=\linewidth]{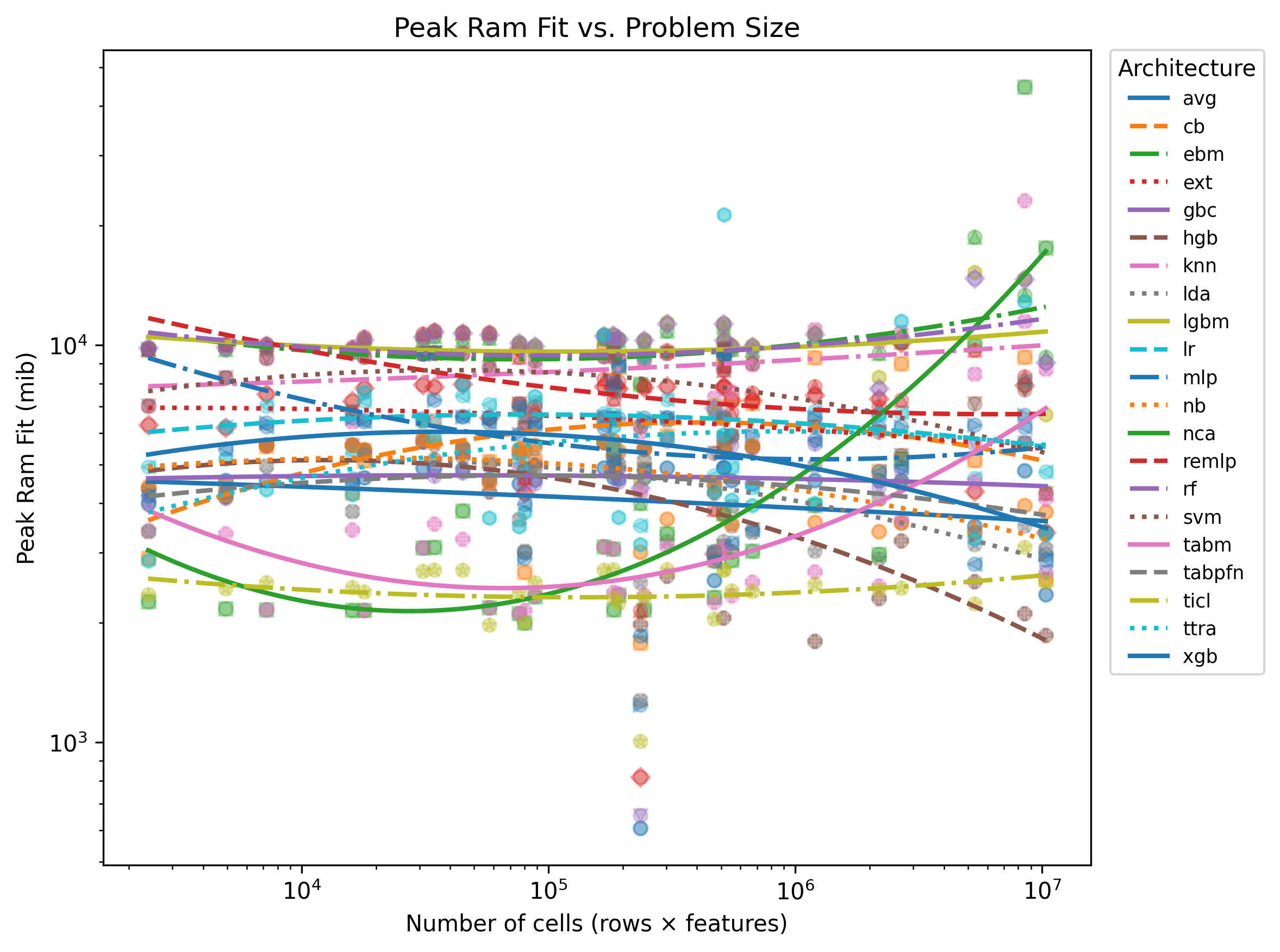}
        \caption{Peak training RAM vs. cells}
    \end{subfigure}
    \caption{Log axes. Lines are interpolated second-order polynomials.}
\end{figure}

These plots shows how a model's peak RAM and consumed CPU seconds during training scales with the size of the dataset. 
We see TABM, NCA and REMLP are indeed the highest training compute consumers. While AVG, NB and KNN consume the least. Peak RAM usage scales gently with size expect for NCA and TABM.

\begin{figure}[H]
    \centering
    \begin{subfigure}{0.48\linewidth}
        \centering
        \includegraphics[width=\linewidth]{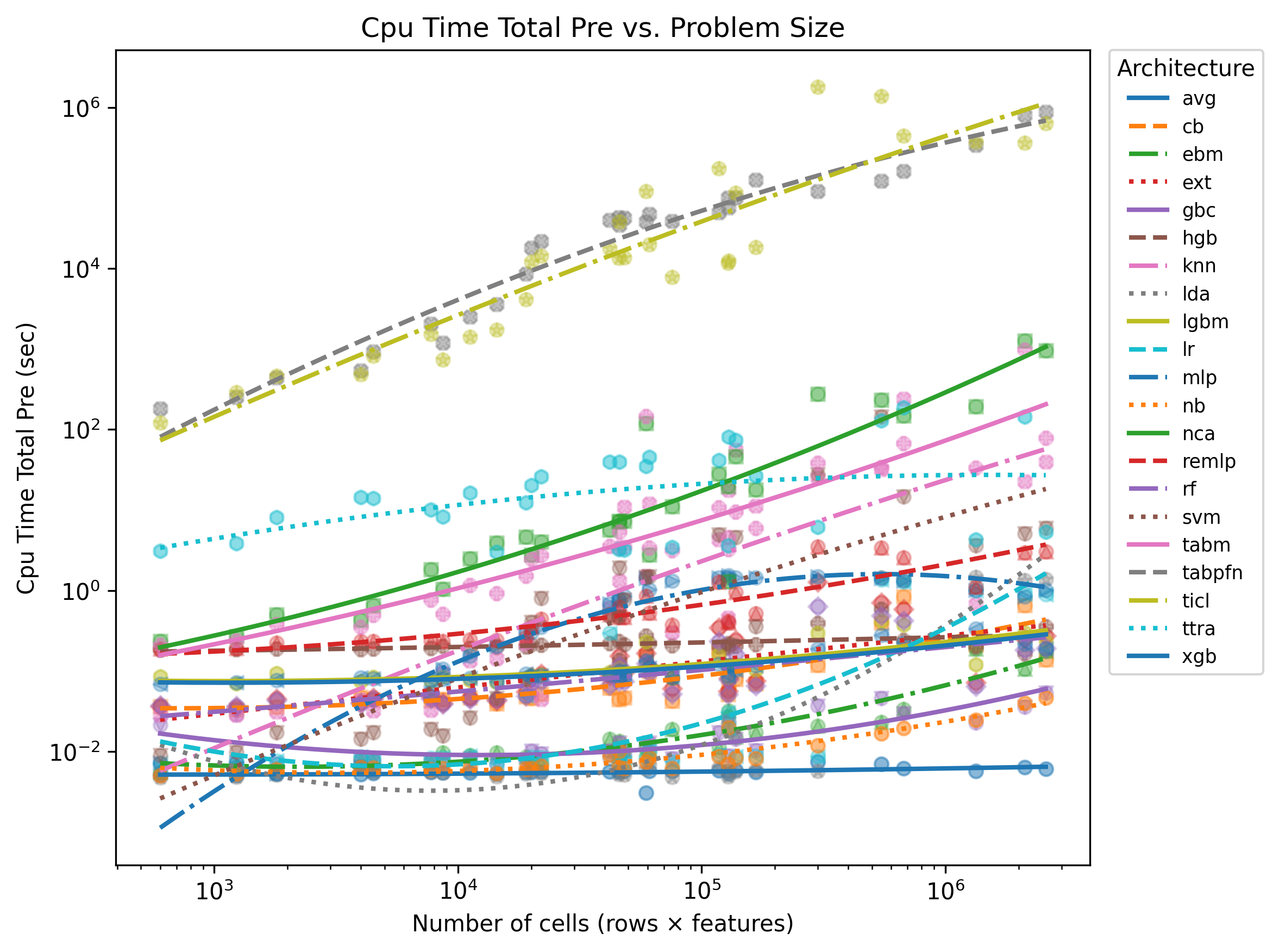}
        \caption{Inference CPU time vs. cells}
    \end{subfigure}
    \hfill
    \begin{subfigure}{0.48\linewidth}
        \centering
        \includegraphics[width=\linewidth]{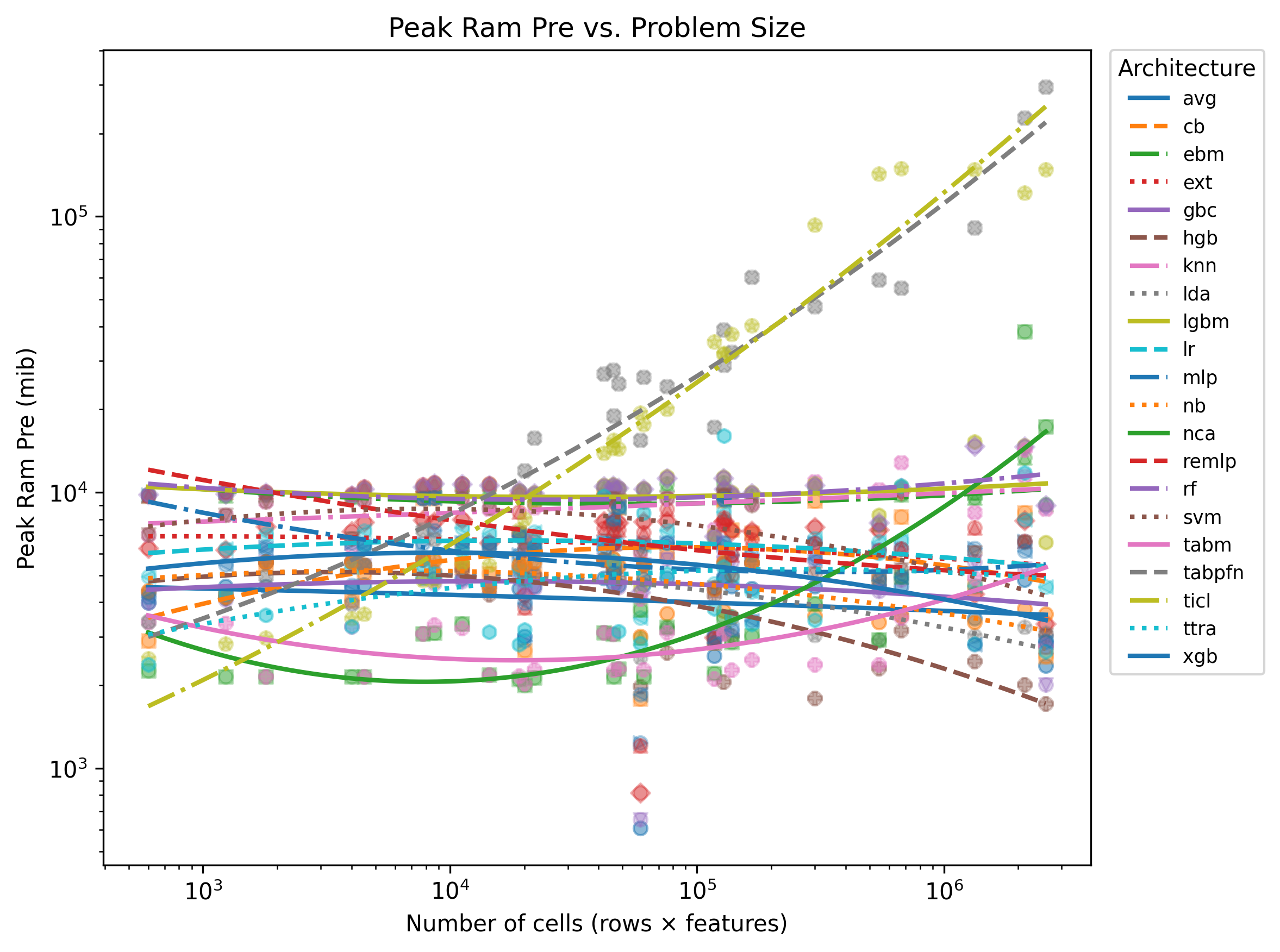}
        \caption{Peak inference RAM vs. cells}
    \end{subfigure}
    \caption{Log axes. Lines are interpolated second-order polynomials.}
\end{figure}

The plot shows how a model's peak RAM and consumed CPU seconds during inference scales with the size of the dataset. We see TICL and TABPFN consumes far more compute then any other model during inference. Their peak RAM usage also scale with dataset size worse then any other model.

\begin{figure}[H]
    \centering
    \includegraphics[width=0.7\linewidth]{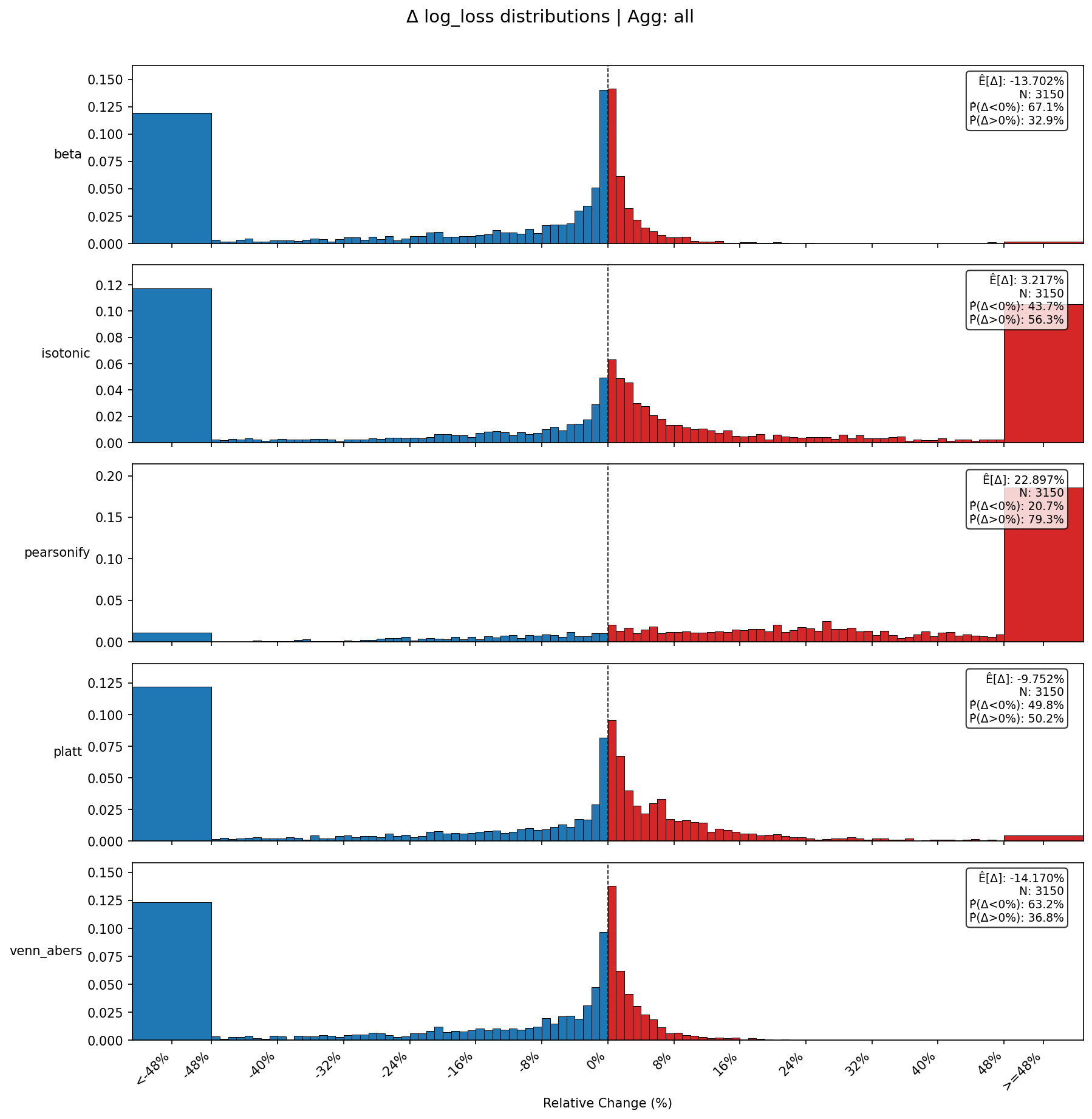}
    \caption{Relative change in log-loss per calibration method across models.}
\end{figure}
In aggregate across all; beta, platt and venn abers are the only calibration methods, among those explored, with an expected improvement in log-loss. Isotonic is expected to slightly increase log-loss while pearsonify markedly increases it. Venn abers is excepted to decrease log-loss the most with $-14.17$\% followed by beta calibration at $-13.7$\% then platt at $-9.75$\%. Beta improves log-loss most frequently at $67.1$\% of instances, followed by venn abers at $63.2$\%. Platt appears to be much of a coin toss at $49.8$\%. But venn abers has less instances of extreme degradation in log-loss, and slightly more instances of extreme improvement.  

\begin{figure}[H]
    \centering
    \includegraphics[width=0.7\linewidth]{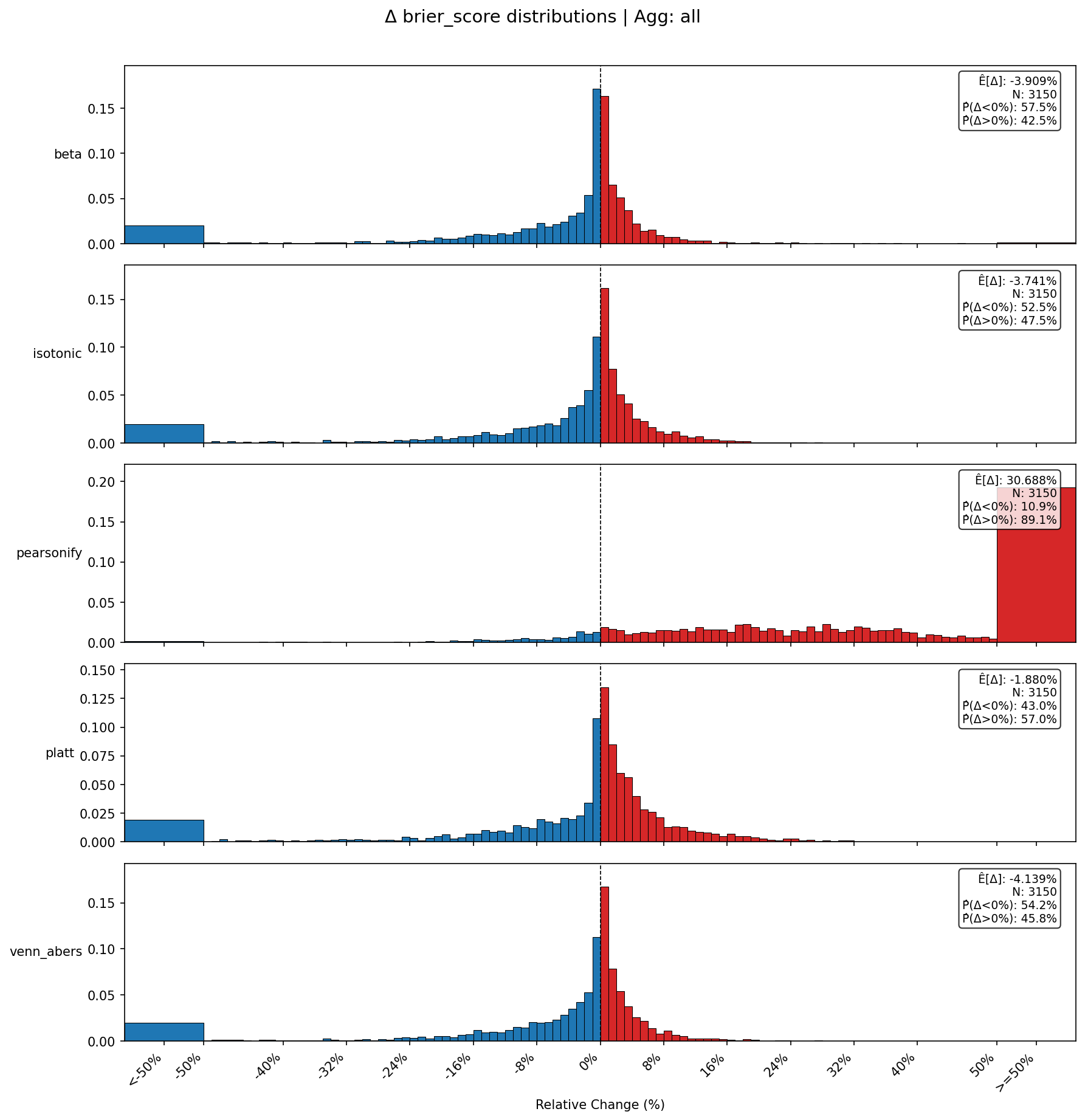}
    \caption{Relative change in brier score per calibration method across models.}
\end{figure}

With regards to brier score all but personify is expected to improve the measure. Venn abers takes first place with $-4.14$\% followed by beta at $-3.91$\%, then isotonic at $-3.74$\%. Beta calibration improves the measure most frequently at $57.5$\% of instances. Venn abers comes second with $54.2$\% then isotonic with $52.5$\%. Again venn abers has fewer instances of extreme degradation. 

\begin{figure}[H]
    \centering
    \includegraphics[width=0.7\linewidth]{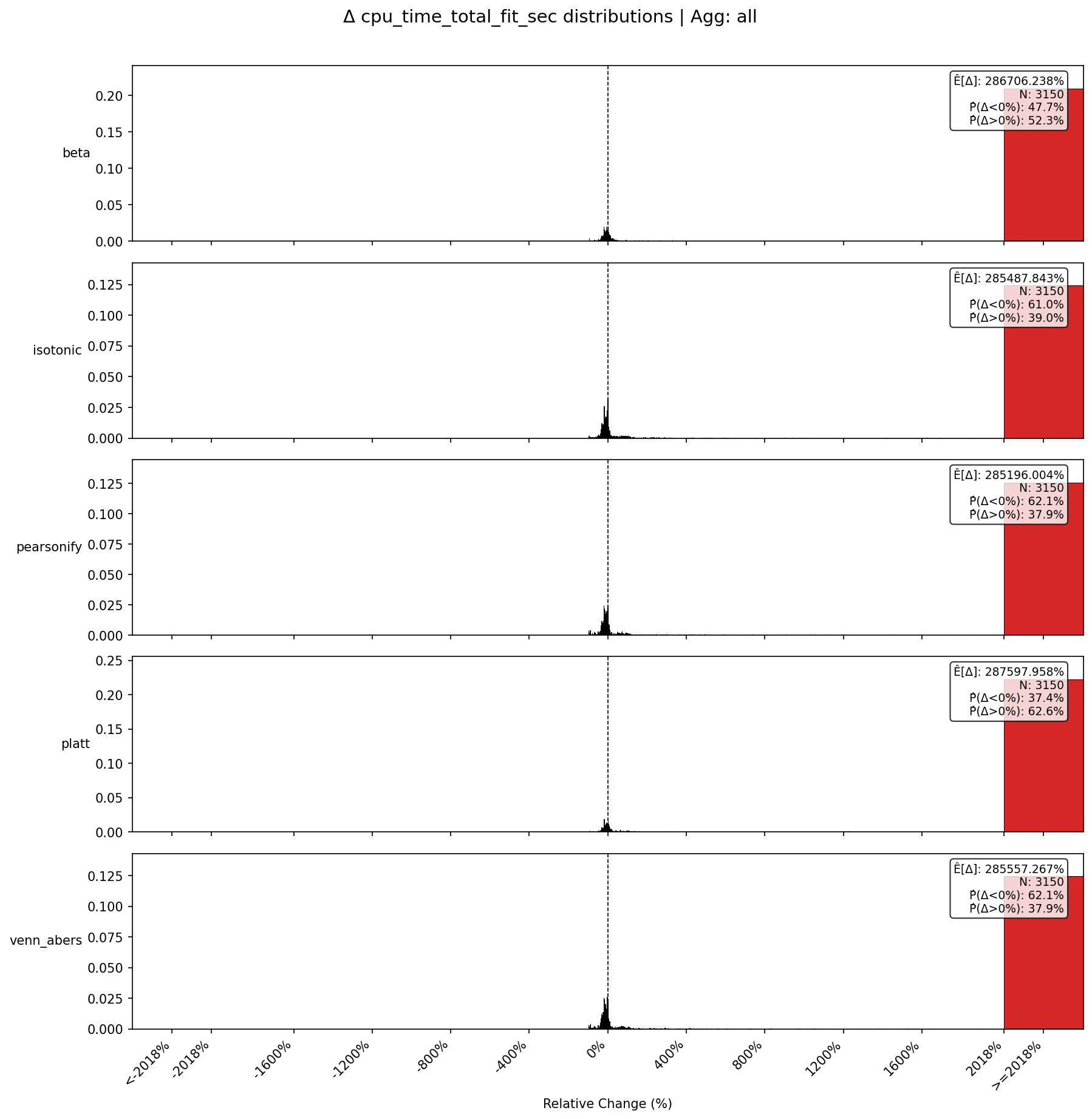}
    \caption{Relative change in CPU seconds during training per calibration method across models.}
\end{figure}
Relative change in CPU costs during training has been heavily tainted by outliers. We'll return to costs later but note for now that these plots are aggregated across all models but there is high inter-model variance. For example, AVG has the lowest training compute costs, and these post-hoc calibration methods significantly increase these costs relative to the base case. Also TICL and TABPFN which have massive inference costs do not lend themselves to post-hoc calibration as this necessitates an inference step on the held-out calibration set during training.

\begin{figure}[H]
    \centering
    \includegraphics[width=0.7\linewidth]{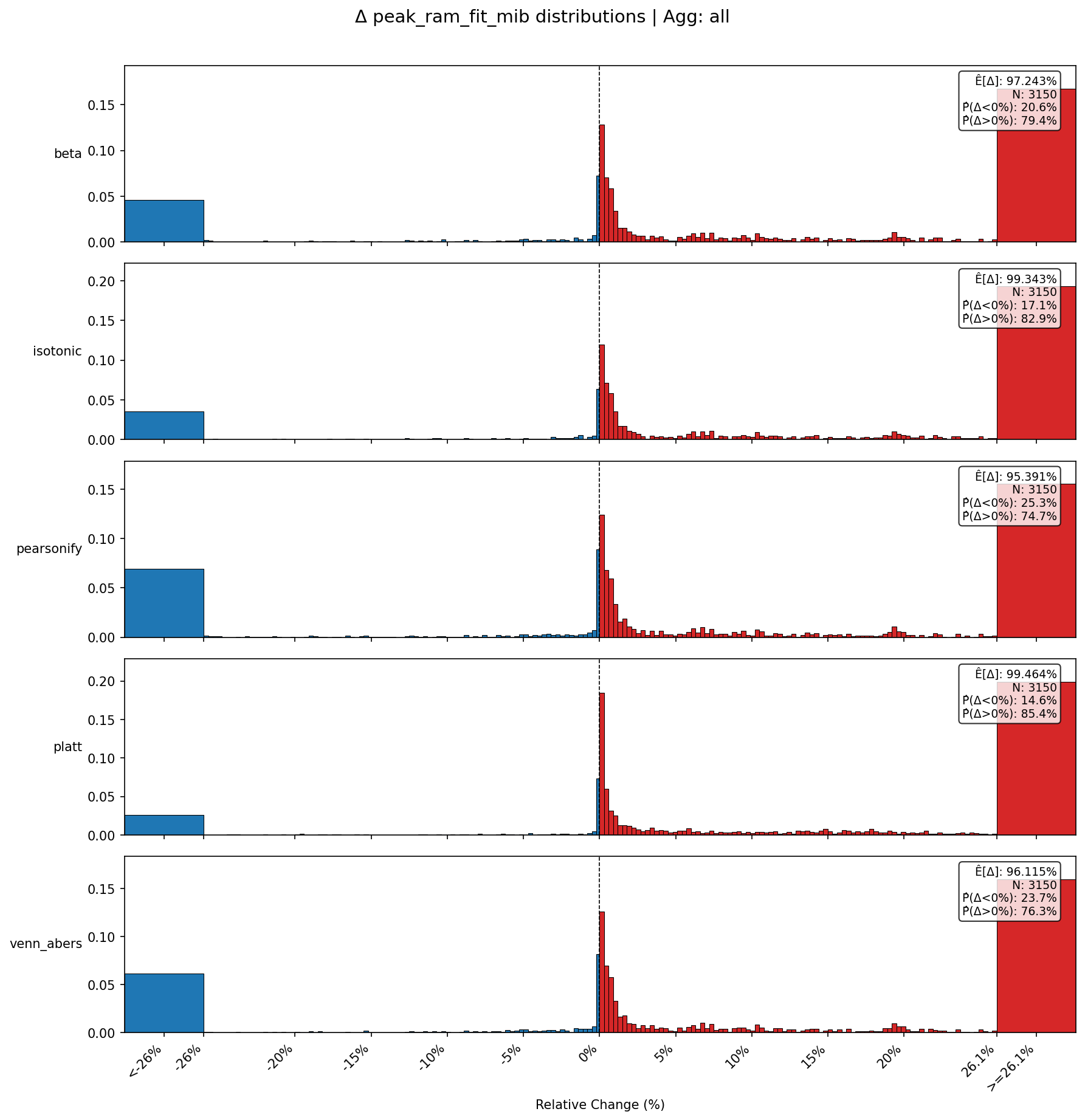}
    \caption{Relative change in peak RAM during training per calibration method across models.}
\end{figure}
All methods are relatively in-line when it comes to their effect on peak RAM during training. They all approximately doubled the maximum RAM consumption during training in expectation.

\begin{figure}[H]
    \centering
    \includegraphics[width=0.7\linewidth]{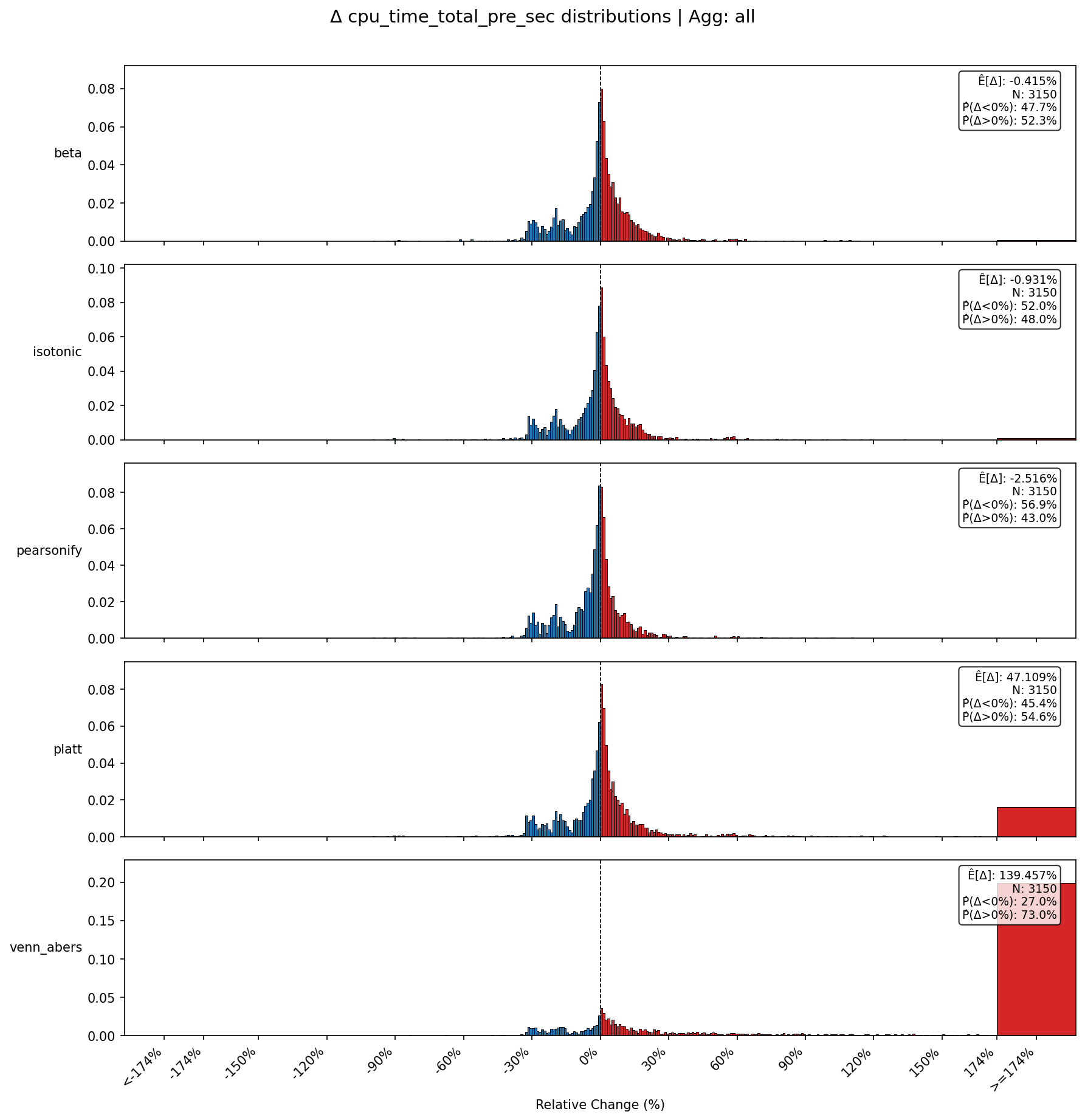}
    \caption{Relative change in CPU seconds during inference per calibration method across models.}
\end{figure}

We see that venn abers and platt increase inference  compute costs significantly, while the remaining methods are expected to decrease it. Venn abers is expected to increase inference time CPU seconds by $139.5$\%  and platt by $47.1$\%. While beta, isotonic and pearsonify are expected to decrease the measure by respectively $-0.42$\%, $-0.931$\% and $-2.5$\%.

\begin{figure}[H]
    \centering
    \includegraphics[width=0.7\linewidth]{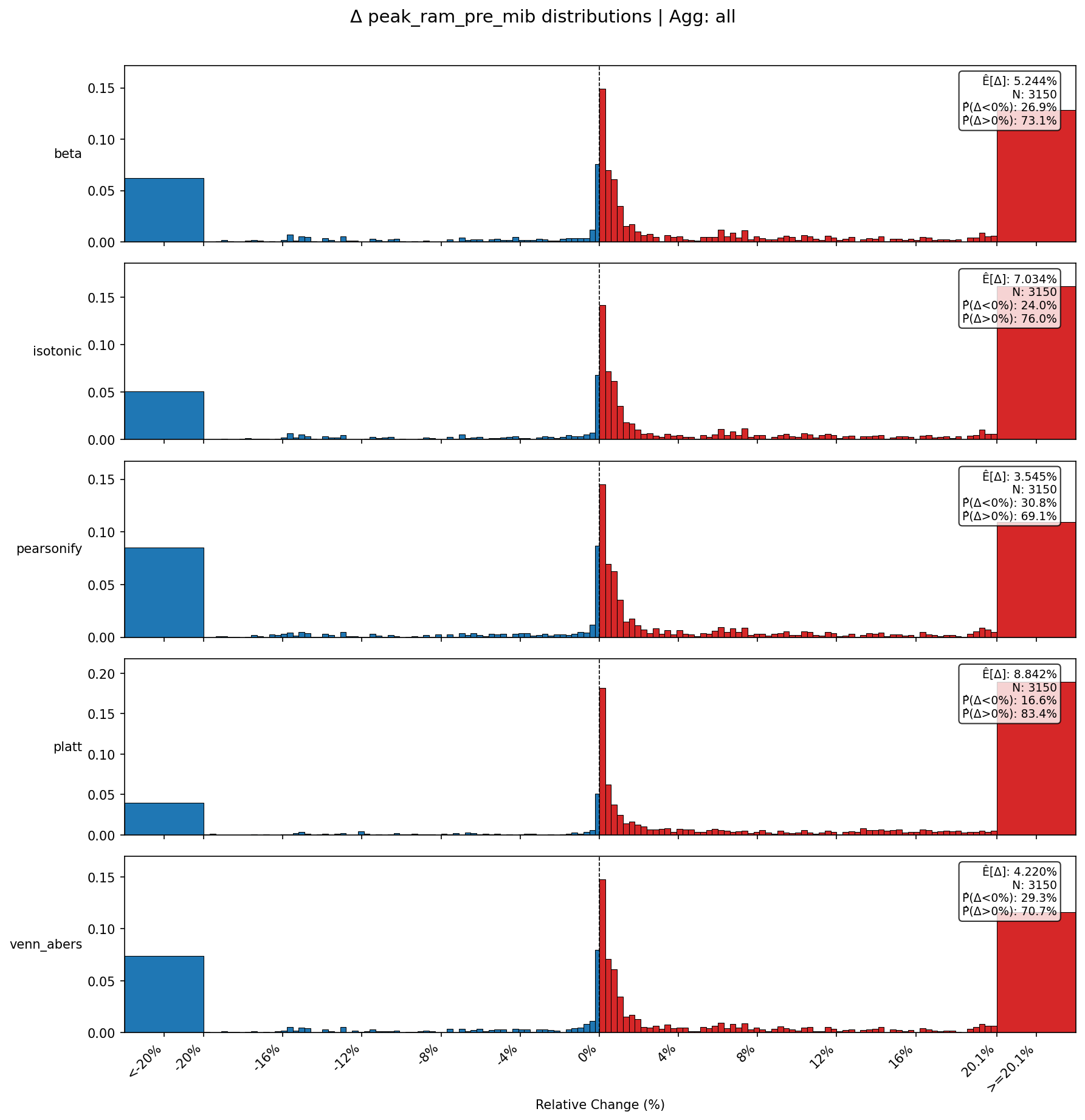}
    \caption{Relative change in peak RAM during inference per calibration method across models.}
\end{figure}
All methods are relatively in-line when it comes to their effect on peak RAM during inference. They all approximately increase the maximum RAM consumption during inference by some percentage points in expectation.

\begin{figure}[H]
    \centering
    \includegraphics[width=0.7\linewidth]{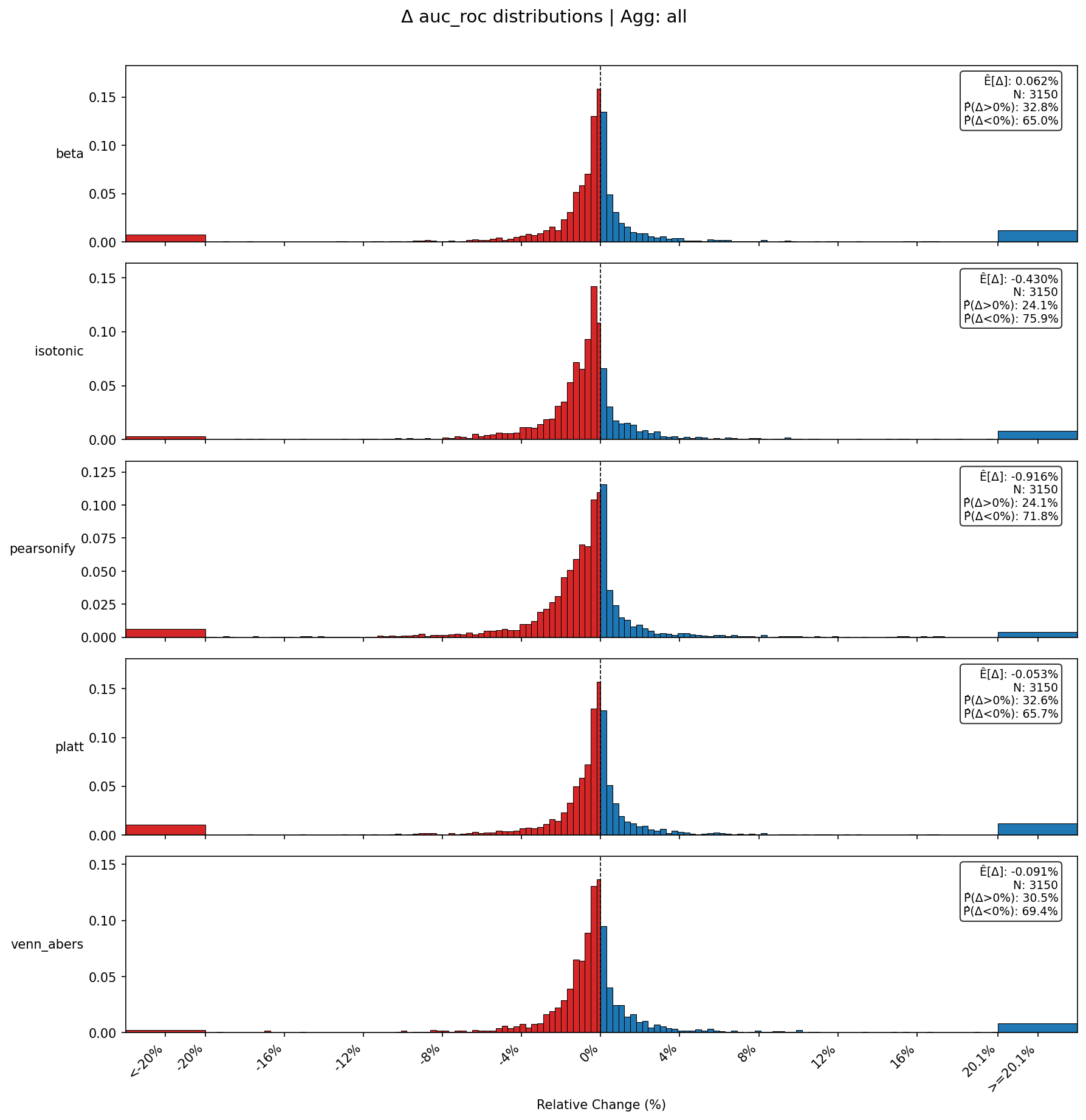}
    \caption{Relative change in AUC ROC per calibration method across models.}
\end{figure}

Only beta is expected to improve AUC ROC by $0.062$\%. All other are expected to degrade the measure by less then a percentage point. Notice that venn abers has less instances of extreme degradation. But all methods degrade the measure more frequently then they improve it. 

\begin{figure}[H]
    \centering
    \includegraphics[width=0.7\linewidth]{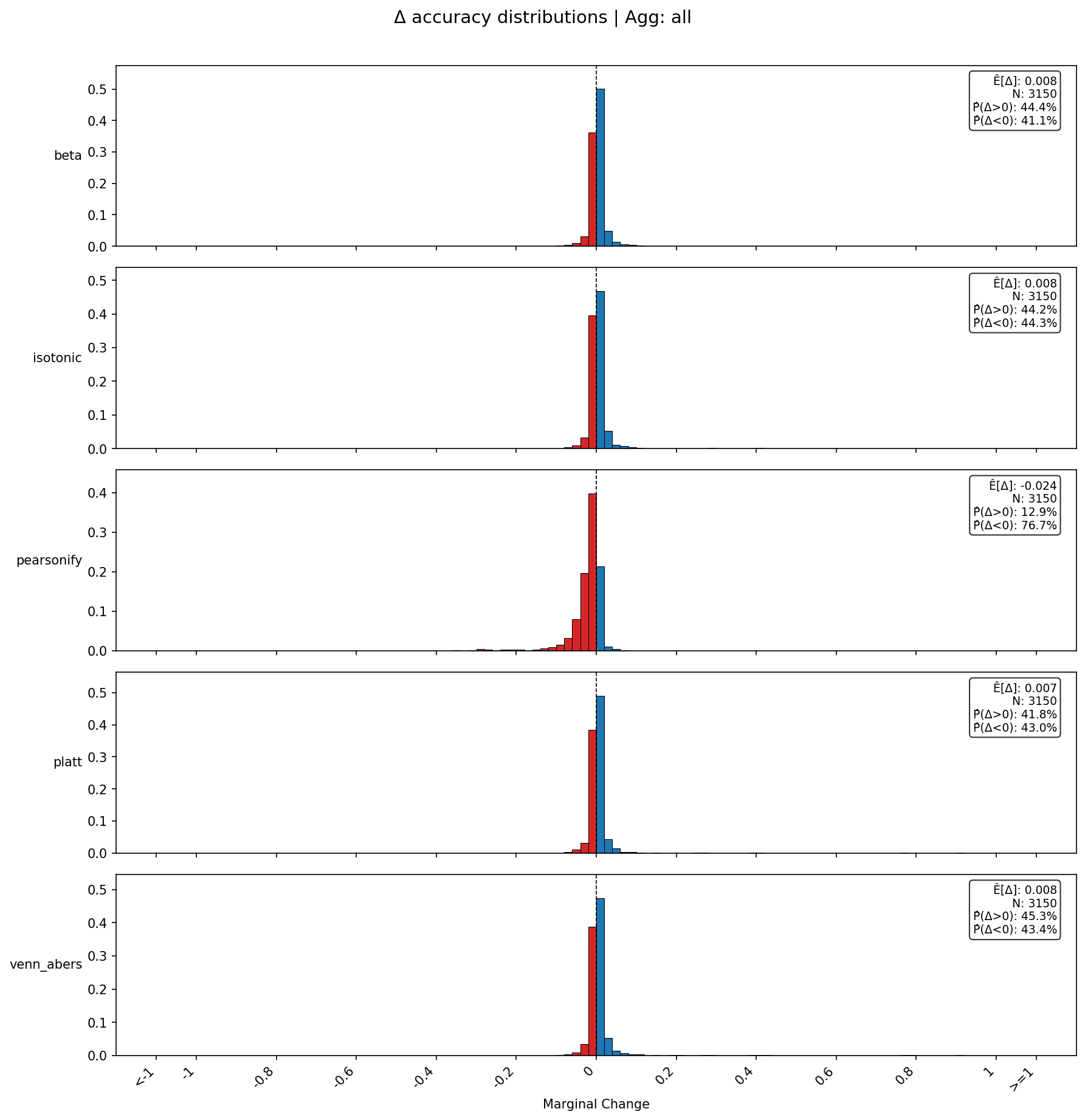}
    \caption{Relative change in accuracy per calibration method across models.}
\end{figure}
All methods but pearsonify is expected to improve accuracy \footnote{Remember, the decision boundary is kept constant at 0.5 in the base case and after a calibration method is used.}. The effect is marginal, and the highest expectation is $0.008$\%.

Moving on to inter-model variance, in appendix \ref{app:plots} we find the expected ranking per architecture by log-loss. We find models which rank ahead of their calibrated counterparts such as TICL, CB and EBM. These models seldom reap benefit from post-hoc calibration in terms of log-loss. However, TABPFN with beta calibration rank ahead of it's uncalibrated counterpart. Note, when applying post-hoc calibration two things change. The base model is trained on a smaller set since a calibration set is held out. Then the model's inferred class score is altered by the calibrator. There is therefore an issue of attributing performance changes between a reduced training set and the calibrator.

\begin{figure}[H]
    \centering
    \begin{subfigure}{0.48\linewidth}
        \centering
        \includegraphics[width=\linewidth]{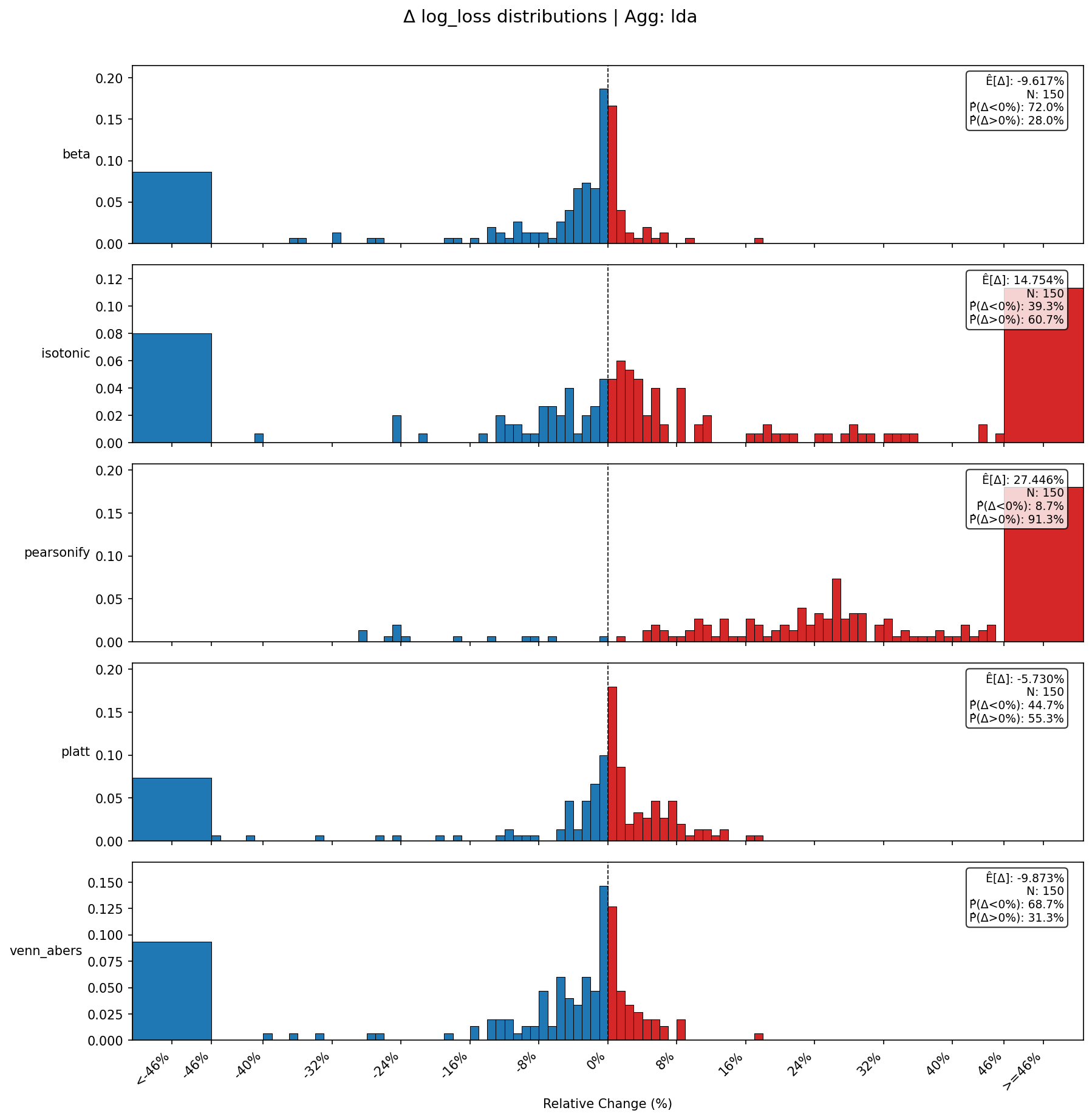}
        \caption{LDA}
    \end{subfigure}
    \hfill
    \begin{subfigure}{0.48\linewidth}
        \centering
        \includegraphics[width=\linewidth]{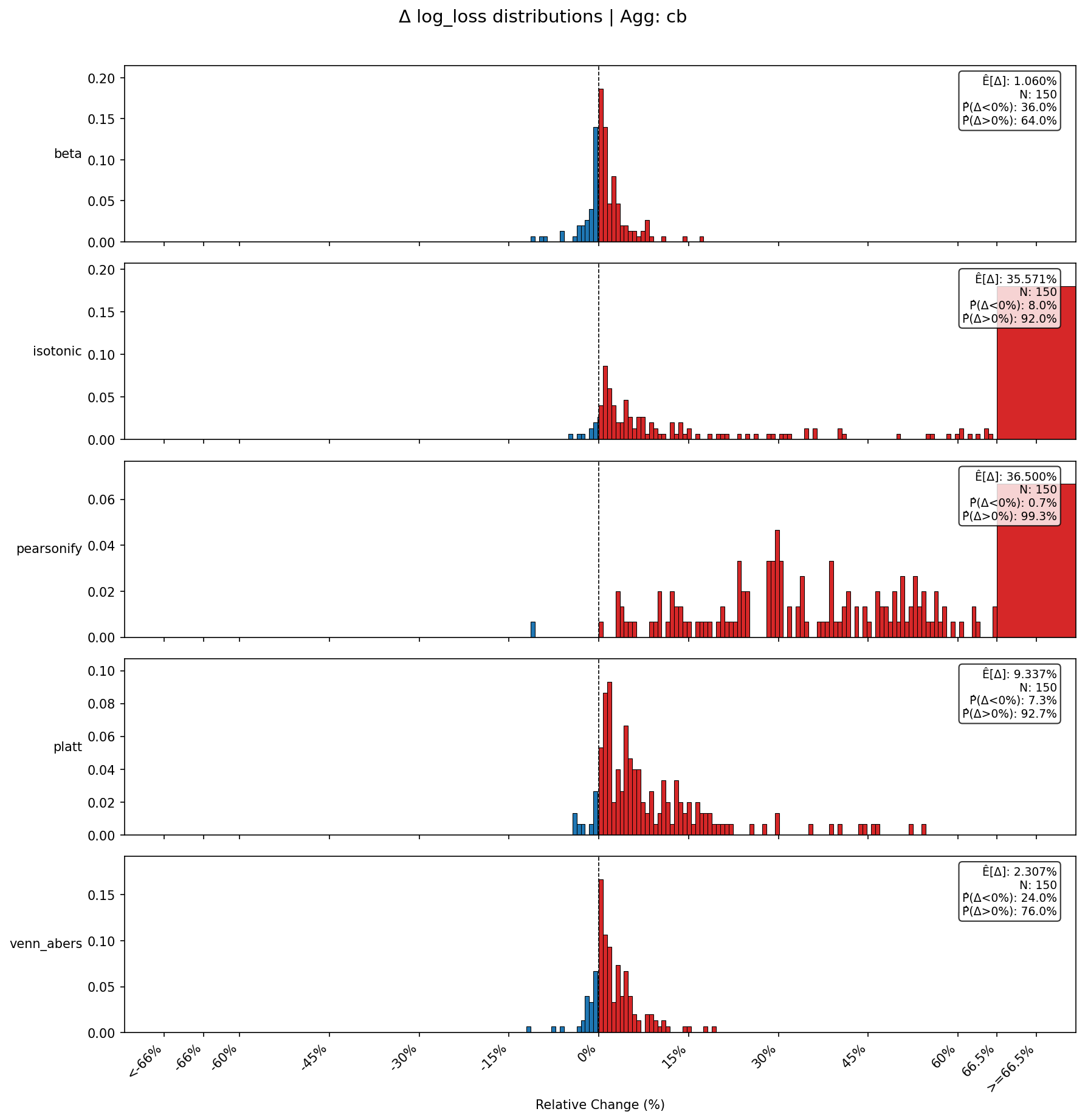}
        \caption{CB}
    \end{subfigure}
    \caption{Relative change in log-loss per calibration method for LDA and CB}
\end{figure}

To further illustrate high inter-model variance we observe the contrast the effect of calibration on log-loss between LDA and CB. We see calibration generally improves log-loss for LDA but degrades it for CB.

\begin{figure}[H]
    \centering
    \begin{subfigure}{0.48\linewidth}
        \centering
        \includegraphics[width=\linewidth]{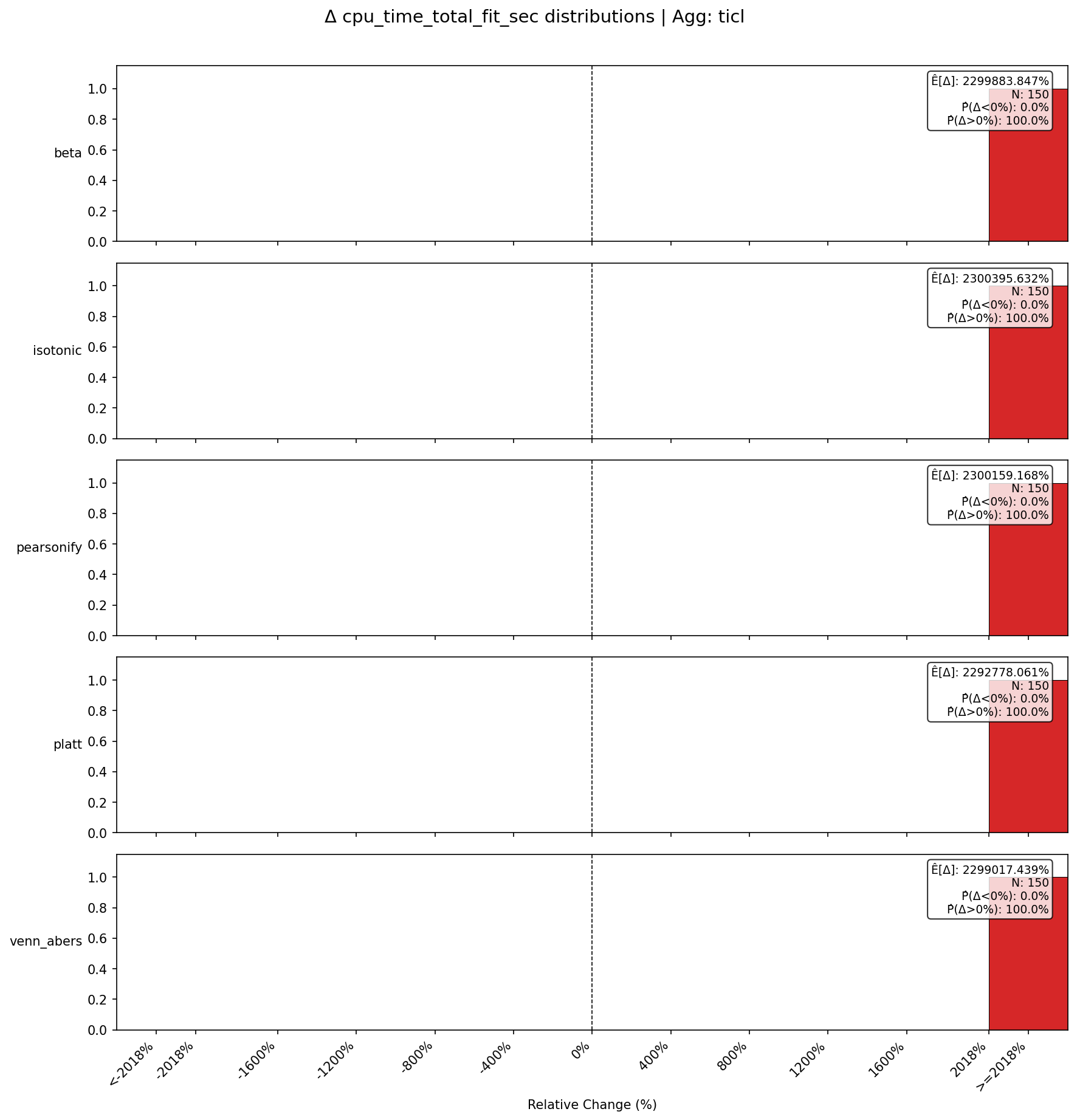}
        \caption{TICL}
    \end{subfigure}
    \hfill
    \begin{subfigure}{0.48\linewidth}
        \centering
        \includegraphics[width=\linewidth]{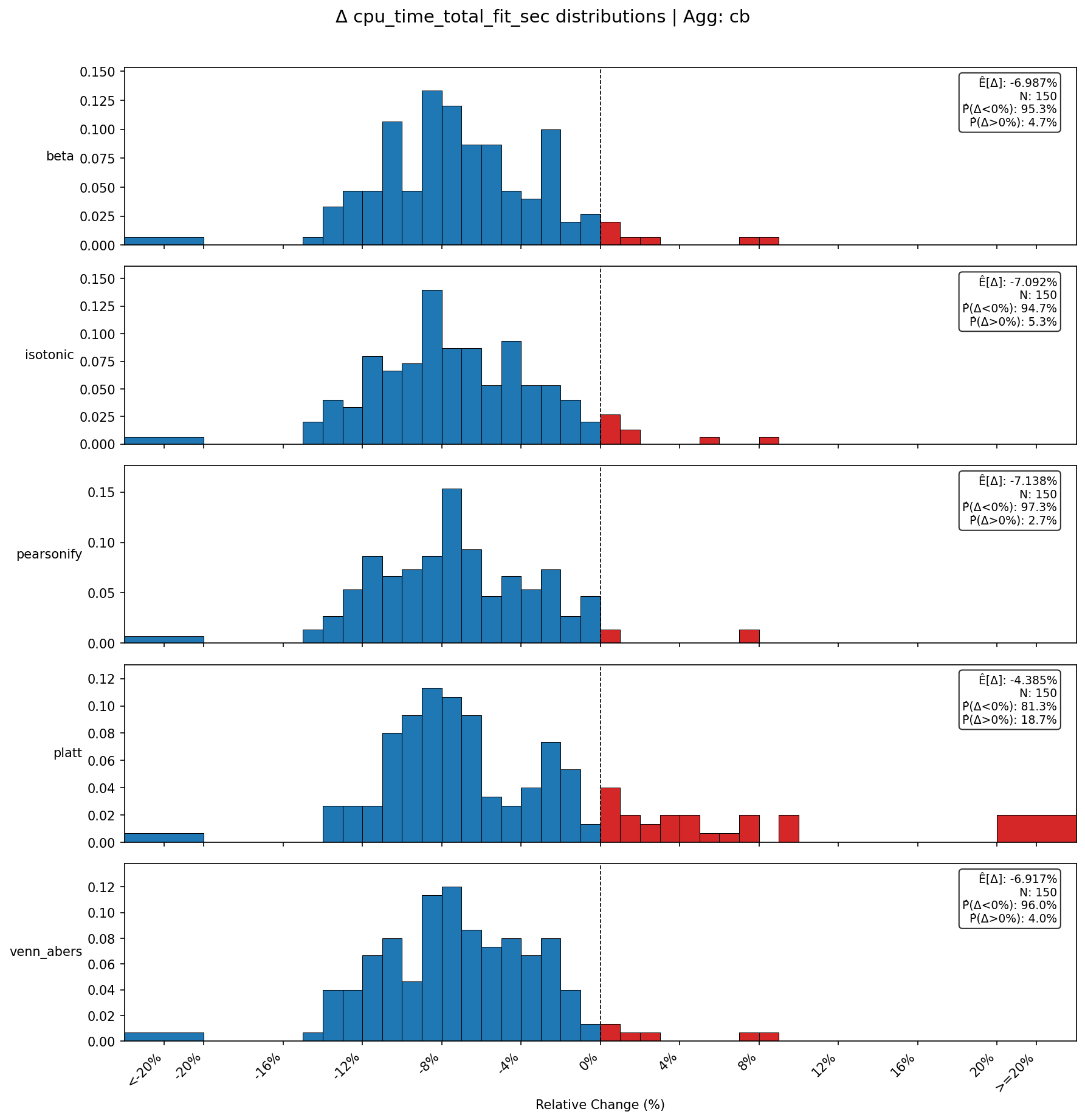}
        \caption{CB}
    \end{subfigure}
    \caption{Relative change in CPU seconds consumed during training per calibration method for TICL and CB}
\end{figure}

We also observe high inter-model variance in the effect on training time compute. TICL's compute consumption is massively degraded by post-hoc calibration due to the inherent inference step. While CB's compute consumption is often improved due to savings from a smaller training set often outweighing the added cost of calibration. 

\section{Conclusion}
Overall, we recommend Beta calibration and Venn--Abers predictors as sensible starting points for practitioners considering post-hoc calibration on tabular binary classification tasks. Across models and datasets, these two methods most consistently improve calibration relative to out-of-the-box baselines. However, post-hoc calibration is not guaranteed to improve performance, particularly for already strong predictors such as TICL, CatBoost, and EBM, where calibration or proper scoring performance can remain unchanged or degrade.

When improvements occur, they are typically accompanied by additional computational cost. Beta calibration and Venn--Abers predictors generally increase compute and memory usage during both training and inference, with the effect usually more pronounced during training. This trade-off raises a practical question of whether available resources might be more effectively allocated to stronger base architectures, additional data, or improved validation and feature engineering rather than post-hoc calibration.

Finally, while calibration quality can change substantially, overall discrimination and classification performance are often preserved, as reflected in relatively stable AUC-ROC and accuracy. These findings highlight that the benefits of post-hoc calibration are strongly dependent on the underlying model, dataset, and resource constraints, and should therefore be evaluated on a case-by-case basis prior to deployment.

\section{Foundation and ideas}
\subsection{Supervised k-class classification}

A supervised k-class classification problem on tabular data is the task of finding the optimal function $f: \mathcal{X} \rightarrow \mathcal{Y}$, given a tabular dataset $T = \{ (\textbf{x}_{i}, y_i) \}_{i=1}^{N} \subset \mathcal{X} \times \mathcal{Y} $. $f$ is an element in the infinite set of functions from $\mathcal{X}$ to $\mathcal{Y}$, $f \in \mathcal{F}(\mathcal{X}, \mathcal{Y})$, where $f(\textbf{x}) = \hat{y} $. Optimum is defined as the extremum of a performance measure over the function space, given a dataset, often formulated as the global minimum of a loss function over $\mathcal{F}(\mathcal{X}, \mathcal{Y})$. \citep{shalev-shwartz_understanding_2014}

$ \mathcal{X} $ is the domain set or the instance space where $\textbf{x} \in \mathcal{X}$ and $ \textbf{x} \in \mathbb{R}^{n} $. $\textbf{x}$ is then a vector with n dimensions. While, $\mathcal{Y}$ is the label set where $y \in \mathcal{Y}$ and $\mathcal{Y} = \{1,2, \cdots,k \}$. In Binary problems, $k=2$ and $\mathcal{Y} = \{0,1\}$. \citep{shalev-shwartz_understanding_2014}

$T$ is assumed under I.I.D. to have been generated by the joint probability distribution $\mathcal{D}$.
Each row $(\textbf{x}_{i}, y_i)$ is then viewed as an independent sample from a multivariate random variable distributed as $\mathcal{D}$, which implies that $\mathcal{D}$ has a functional form $ d $ which is the true probability density function. \citep{shaker_random_2025}

The I.I.D. assumption for a dataset $T$ is not guaranteed. When we generate synthetic data by sampling a multivariate random variable, we can enforce independence and identical distribution, or deliberately violate it. In contrast, with real‐world data we lack access to the true generative process, so we cannot  confirm those assumptions, though we can justify them. As a result, you must always justify an I.I.D. assumption and be conscious of potential distributional shifts in the future. \citep{gardner_benchmarking_2023}

 $f$ seldom maps onto $\mathcal{Y}$ "directly", and a common component function has the standard k-1 dimensional probability simplex as codomain:
 \begin{gather*}
 \textbf{f}: \mathcal{X} \rightarrow \Delta^{k} \\
  \textbf{f}(\textbf{x}) = \hat{\textbf{y}} \\   
  \hat{\textbf{y}} = [p_1, \ p_2, \ \cdots, \ p_k]
 \end{gather*}
 Note, in binary scenarios many methods operate on the positive class' score $p_1$, instead of the vector $\hat{\textbf{y}}$ since the score of the negative class $p_0$ is always $1 - p_1$ and the class' encoding is $0$. \citep{silva_filho_classifier_2023}  
 
 The highest class score $p_j$ or another function on $\Delta^{k}$ determines $\hat{y}$ (Bayes classifier), such that:
 \begin{gather*}
  c:\Delta^{k} \rightarrow \mathcal{Y} \\   
  f = c \ \circ \ \textbf{f}
 \end{gather*}
 
In industry applications, $c$ ought to maximize expected value \citep{flores_consequentialist_2025}.
$\hat{\textbf{y}}$ is either "directly" the learner's output or derived from it's outputs using a normalization function, $s:\mathcal{O} \rightarrow \Delta^{k}$ (e.g. softmax). $\mathcal{O}$ , the output space of the learner, is commonly called logit space for neural networks, and the normalization function the normalization layer. While for SVMs and margin classifiers it takes the name margin space or discrimnant-function space. 

\subsection{Architectures}
We explicitly distinguish between a \textbf{learner / model} and an \textbf{architecture}. By \emph{learner} or \emph{model} we mean a self-contained procedure (e.g., \href{https://catboost.ai/docs/en/concepts/python-reference_catboostclassifier}{CatBoostClassifier}) that, given training data \(T\), produces a function \(\mathbf{f}\). The learner excludes pre-/post-processing, feature selection, hyperparameter search, sampling strategies, ensembling, etc., unless those steps are contained within the procedure itself. An \textbf{architecture} is the complete composition, the end-to-end process that generates \(\mathbf{f}\) from raw input \(T\), explicitly including learners together with pre-processing, post-processing, tuning, ensembling, and other procedural components. 

Importantly, the process that yields $\textbf{f}$ need not begin from scratch, an architecture can incorporate priors or artifacts (for example, via pre-training, distillation, transfer learning, meta-learning, continual learning, or by initializing from previously trained components) as part of the end-to-end generation. Beyond continual learning, we might also envision mechanisms that enable the effective transfer of priors and artifacts across different architectures, such that any newly instantiated model can inherit or initialize from the accumulated "knowledge" of its predecessors. This would allow us to capitalize on the compute already expended during prior training, in a manner reminiscent of Isaac Newton’s famous remark: “If I have seen further, it is by standing on the shoulders of giants.” In contrast, many models trained purely from scratch begin as blank slates, and once they complete their task, their acquired "experience" is not natively passed forward. To overcome this, one could imagine a distributed network of models in which practitioners retrieve priors from a shared repository to initialize their architectures, and, upon completing their training or generative process, contribute their own learned experience back to this network. In doing so, the community could accumulate and transmit knowledge across generations of architectures, gradually compounding collective progress. Parallels can be drawn to the work of \citep{cotton_microprediction_2022}.

An architecture is therefore compositional: it integrates all components. For a real-world example, see Amazon’s Rate Card Transformer \citep{sreekar_unveiling_2023}; conceptually one can for example write:
\[\textbf{f} = Platt \ scaling \ \circ \ CatBoost \ \circ \cdots \circ \ Imputation \] 
The architecture is a higher-level abstraction than a single learner, although in some cases the learner \emph{is} the entire architecture (e.g., K-NN applied to a clean numeric table). But if you impute missing values before K-NN, then \(\text{K-NN} \circ\text{Imputation}\) is the architecture. Thus the learner/model becomes a component within a broader architecture, while the architecture is the top-level abstraction.

We've observed that components are often chosen or optimized in isolation (pre-processing, feature engineering, model choice considered separately), that modular approach can miss important interaction effects. Ideally the full composition \(f\) should be treated as a single, unified function for optimization: joint optimization across stages better captures cross-stage dependencies and can approach globally optimal solutions instead of settling for a chain of locally optimal choices.

The architectural abstraction can be extended beyond the modeling pipeline: upstream to \textbf{data provenance}, downstream to \textbf{decision making}, vertically/horizontally to the \textbf{computer platform}.

\begin{itemize}
  \item \textbf{Data provenance} covers the data collection mechanism, storage and retrieval systems, and raw signal processing. A medical diagnosis architecture might therefore begin at the sensor level:
  \[
  \mathbf{f} = \text{K-NN}\ \circ\ \text{Imputation}\ \circ\ \cdots\ \circ\ \text{ECG Denoising}\ \circ\ \text{Electrode Sampling}.
  \]
  \item \textbf{Decision making} covers the consequences or actions following a classification. Essentially, how they are consumed by business logic or action modules (alerts, treatment recommendations, automated control). For example, the final decision function $g$ could be written as
  \begin{gather*}
  f = \text{Diagnose highest-likelihood disease}\ \circ\ \mathbf{f} \\
  g =  \text{Treat said disease} \ \circ \  f
  \end{gather*}
  
  \item \textbf{Computer platform} 
    denotes the full stack that executes an architecture, from silicon to system software and network: compute devices (CPUs, GPUs, TPUs, FPGAs, ASICs, and emerging photonic platforms such as \href{https://lightmatter.co/}{Lightmatter}), instruction sets and microarchitectures, memory and storage hierarchies, NICs and fabric, OS/driver/runtime layers, utility libraries, protocols, compilers and optimizers (e.g., XLA, LLVM), container/VM layers, and deployment topology (single node, multi-node cluster, edge, cloud). This dimension recognizes that architectural behavior and performance are tightly coupled to hardware paradigms and their software implementations. For example:
  \begin{itemize}
    \item \textbf{Parallelization efficiency is algorithm- and platform-dependent.} Embarrassingly parallel designs (e.g., many independent ensemble members) can scale near-linearly on distributed topologies because they require little inter-process communication, while tightly coupled workloads need high-bandwidth, low-latency interconnects and communication-aware partitioning.
    \item \textbf{Hardware acceleration targets specific computations and precisions.} Dense linear algebra and matrix-multiply–heavy workloads frequently achieve large (sometimes orders-of-magnitude, under favorable conditions) speedups on accelerators (GPUs/TPUs) versus CPUs, particularly with large batch sizes, reduced precision (BF16/FP16/INT8), optimized kernels, and compilation/fusion in the software stack (BLAS/CUDA/XLA/etc.). Realized gains depend strongly on batch size, memory bandwidth, numerical precision, kernel fusion, and the software stack.
    \item \textbf{Implementation and runtime artifacts materially affect latency and throughput.} Ahead-of-time compiled implementations (C++, Rust) or well-tuned native libraries remove interpreter overhead and enable low-level memory/control optimizations, commonly yielding substantially lower latency than for example Python-level implementations; the exact factor varies with workload, I/O, and parallelism characteristics.
  \end{itemize}
\end{itemize}
The architecture thus recursively abstracts all steps from bio-electric signal acquisition to clinical treatment and the platform on which it operates.

This holistic, end-to-end view extends beyond our formulation of a supervised k-class classification problem on tabular data. With this perspective we instead search for optimum (e.g, extremum of value) over the space of architectures:
\begin{gather*}
    \mathcal{A} = \{ \mathcal{F}(\mathcal{W}_{t}^{s}, \mathcal{I}_{t + \delta}) \times \mathcal{P} \ | \ c(\textit{f},\textit{p}) \leq \mathcal{C}, \  \textit{f}\in \mathcal{F}, \ \textit{p}\in \mathcal{P}\ \}\\
    \mathcal{A} = \text{Architectures} \\
    \mathcal{W} = \text{Accumulated World State in space s at time t} \\ \
    \mathcal{I} = \text{Action space at time t} + \delta \\
    \mathcal{P} = \text{Computer platform} \\
    c(\textit{f},\textit{p}) \leq \mathcal{C} = \text{Operates within constraints/requirements}
\end{gather*}
Over time you record observations or experience to generate $\textit{f}$.

\subsection{Datasets}

We require a dataset suite;  
\begin{gather*}
    \{ T_l \mid l,L \in \mathbb{N}, \ l \le L\}
\end{gather*}
that sufficiently represent the diversity of datasets encountered in practice. Ideally, this collection should span plausible, real-world characteristics that a practitioner is likely to observe.

Mathematically, the space of all possible datasets is infinite, as each $d_l$ is a member of the infinite-dimensional space of probability density functions. To make this space tractable, we identify key dimensions along which datasets typically vary, and construct a finite, representative benchmark suite by selecting datasets that capture variation along these dimensions.

\subsubsection{Synthetic}
For synthetic datasets, we can precisely control the data-generating process, allowing us to systematically explore a wide range of scenarios with diverse statistical properties.

To generate a synthetic I.I.D. dataset $T_l$ we:   
\begin{enumerate}
    \item Draw $N_l$ samples from a multivariate random variable distributed as $d_l$, which returns $\mathcal{T}_l$
    \item Apply a transformation $G$ on $\mathcal{T}_l$ such that $G(\mathcal{T}_l) = T_l$  
\end{enumerate}

To influence the characteristics of the resulting dataset $T_l$, one must therefore vary the underlying density function $d_l$, the transformation function $G$, or both.

Key dimensions along which $d_l$ may vary include:

\begin{enumerate}
    \item \textbf{Continuity} \newline
    The distribution may be continuous (e.g., Gaussian) or discrete (e.g., categorical, Poisson).
    
    \item \textbf{Functional Form} \newline
    The distribution may exhibit different structural patterns such as linear, polynomial, exponential, trigonometric, or piecewise forms governing the relationships among variables.
    
    \item \textbf{Tail Behavior} \newline
    The density may feature heavy-tailed properties, influencing the prevalence of outliers and extreme values.
    
    \item \textbf{Heteroskedasticity} \newline
    The variance of the error term may depend on the values of the independent variables, affecting the spread of the data.
    
    \item \textbf{Multicollinearity} \newline
    The distribution may induce strong correlations among independent variables.
    
    \item \textbf{Endogeneity} \newline
    The data-generating process may involve correlations between explanatory variables and the error term.
\end{enumerate}

In practice, the density function $d_l$ can be constructed using multivariate parametric families (e.g., Gaussian, Student-\textit{t}), multi-modal distributions (e.g., Gaussian mixtures), copula-based constructions that combine arbitrary marginal distributions, and nonparametric kernel-based densities. These methods provide control over the shape and complexity of the underlying distribution, including the ability to capture complex interactions, skewness, or tail behavior. 

The transformation function $G$ can also be varied to simulate real-world data imperfections and structure. Key dimensions along which $G$ may vary include:

\begin{enumerate}
    \item \textbf{Noise} \newline
    Noise can be introduced in several forms such as missing values, corrupted entries, or artificial interventions. Thereby degrading signal quality and simulating common data quality issues.
    
    \item \textbf{Feature Alteration} \newline
    Irrelevant features may be introduced, or relevant features may be removed, thereby altering the feature set in ways that challenge a model’s ability to generalize.
    
    \item \textbf{Sporadic Relationships} \newline
    Spurious correlations between input variables and the target can be embedded to mimic coincidental or unstable associations often encountered in practice.
\end{enumerate}

Multivariate random variables may model the target variable $\mathcal{Y}$ as continuous. To accommodate classification models, $G$ may discretize the target via binning into $k$ ordinal classes, effectively transforming a regression task into a classification one.

Several prior works have focused on the design of synthetic benchmark datasets:

\begin{itemize}
    \item \textbf{\citep{orzechowski_generative_2021}} introduced the DIverse and GENerative ML Benchmark (DIGEN), which consists of 40 closed-form functions mapping continuous inputs to binary outputs. These functions were discovered through a heuristic optimization procedure aimed at maximizing diversity in model performance. The benchmark spans regimes that are linear, nonlinear, interaction-driven, and noise-dominated.
    
    \item \textbf{\citep{friedman_regularized_1989}} proposed six synthetic classification scenarios. These are all multivariate normal distributions with varying mean vectors and covariance matrix structures. 
    
    \item \textbf{\citep{breiman_arcing_1998}} designed  four synthetic classification problems, namely: waveform, twonorm, threenorm, and ringnorm. Waveform is generated from a mixture model involving Gaussian noise and base signals, while two-, three- and ringnorm are constructed from multivariate normal distributions with varying means and covariances structures. Threenorm includes a mixture of Gaussians within one class, and ringnorm involves classes with different variances, making it effectively a mixture scenario as well.
    
\end{itemize}

Since our focus is on calibration, an essential requirement for synthetic data is that the true conditional class probabilities are known. This enables precise evaluation of the extent to which model-predicted probabilities align with the true data-generating process, thereby supporting rigorous benchmarking of calibration performance. Notice, that the globally optimal $\textbf{f}$ is the true conditional probability function dervied from $d_l$.

\subsubsection{Real}
Unlike synthetic data, real-world datasets are not under our control with respect to their data-generating processes. As a result, we must identify and collect datasets that span a variety of meaningful characteristics.

Real-world tabular datasets can be differentiated along both \textit{qualitative} and \textit{quantitative} dimensions \citep{anonymous_comprehensive_nodate}:

\textbf{Qualitative characteristics} include:
\begin{itemize}
    \item \textbf{Domain origin}: The domain from which the dataset arises such as finance, healthcare, biology, or social sciences.
    \item \textbf{Modality}: The fundamental form in which real-world phenomena were measured and digitally represented. Modality describes how observations from the physical world, such as physiological signals, visual scenes, textual communications, or behavioral interactions were captured through specific measurement processes and encoded into digital formats. Examples include:
\begin{itemize}
    \item Raw Sensor Data: ECG traces, camera images, microphone audio recordings
    \item Direct Observations: Manually entered fields (patient age, product categories, survey responses)
    \item System Logs: Transactional records, server logs, clickstream data, audit trails
\end{itemize}
\end{itemize}

Modality fundamentally determines the data's intrinsic characteristics including measurement units, sampling frequency, spatial/temporal resolution, file formats, noise patterns, and typical missingness mechanisms. These properties in turn dictate the appropriate modeling approaches.

Crucially for tabular models: Regardless of original modality, tabular models inherently operate on numeric matrices and therefore \emph{require} the original modality to be projected into a numeric tabular representation through modality specific feature extraction/engineering:
\begin{itemize}
    \item Sensor/Image/Audio Data: Extracting summary statistics, spectral features or pre-trained embeddings. Such as \citep{dosovitskiy_vision_2020}.
    \item Text Data: Text extracted via OCR: computing bag-of-words representations, TF-IDF scores, contextualized embeddings or embeddings from NLP models.
    \item Logs/Transactions: Aggregating counts, calculating temporal deltas, constructing user/item profiles
    \item Categorical Data: Applying encodings such as one-hot, target encoding, or learned embeddings
\end{itemize}

Note, that there are models natively capable of operating on multiple modalities such as those found in pytorch-widedeep \citep{zaurin_pytorch-widedeep_2023}. Bypassing a tabular representation entirely may also result in higher performance \citep{akhauri_performance_2025}.

\textbf{Quantitative characteristics} include:
\begin{itemize}
    \item \textbf{Class imbalance}: The distribution of class labels may be skewed.
    \item \textbf{Noise and data quality}: Missing values, label noise, or inconsistencies in feature encoding.
    \item \textbf{Data scale and types}: Datasets may vary in the number of instances, dimensionality, and the nature of features (e.g., continuous, categorical, ordinal).
    \item \textbf{Difficulty}: Certain datasets consistently produce lower predictive performance across models, suggesting inherent complexity or noisiness. These are often referred to as “hard” datasets.
\end{itemize}

Distributional shifts such as temporal or domain shifts can also affect calibration, though we omit these in this study to maintain an I.I.D. assumption throughout.

Quantitative properties can also be systematically extracted through automated tools such as the metadata extraction framework for tabular data called pymfe \citep{alcobaca_mfe_2020}. Notably, the characteristics listed above overlap with the design dimensions considered for synthetic datasets.

A number of real-world dataset benchmarks have been developed for general-purpose benchmarking:

\begin{itemize}
    \item \textbf{OpenML-CC18} \citep{bischl_openml_2021}: A curated collection of 72 classification datasets from varied domains such as finance, medicine and manufacturing.
    \item \textbf{TabZilla-Hard Benchmark} \citep{mcelfresh_when_2024}: A set of 36 tabular datasets identified as consistently challenging across a range of machine learning models.
    \item \textbf{\citep{grinsztajn_why_2022}}: A benchmark of 45 datasets from varied domains sourced primarily from OpenML.
    \item \textbf{AutoML Benchmark} \citep{gijsbers_amlb_2023}: Introduced the benchmark AMLB, a comprehensive suite of datasets from OpenML. The classification suite include 71 classification tasks from various domains.
\end{itemize}

Although these datasets are widely used, recent critiques, such as those presented by \citep{tschalzev_unreflected_2025} highlight concerns primarily related to methodological rigor and experimental design, and some flaws in the suites themselves. Consequently, while these benchmarks remain valuable resources, their use should be accompanied by thoughtful dataset selection and appropriate methodological design.

Luckily,  \citep{erickson_tabarena_2025} introduced "Tabarena-v0.1", an open-source and maintained benchmark which includes a suite of 38 versioned, curated tabular classification tasks. These were generated by filtering 1053 datasets from 13 prior benchmark studies (Including all the aforementioned suites). The suite contains datasets from domains such as finance, chemistry, physics, healthcare, and marketing. The filters applied to the original datasets include:

\begin{itemize}
    \item \textbf{Unique datasets}:
    No duplication of datasets.
    \item \textbf{IID Tabular Data}:
    The datasets must be I.I.D.
    \item \textbf{Tabular modality}:
    Exclude datasets which originate from a modality/application where tabular models aren't among the primary architectures.  
    \item \textbf{Real Random Distribution}:
    The dataset must originate from a real random process. No synthethic data. 
    \item \textbf{Predictive Machine Learning Task}:
    Exclude datasets which do not stem from a predictive machine learning task. 
    \item \textbf{Size Limit}:
    Exclude datasets with less then 500 or more then 250 000 samples.
    \item \textbf{Data Quality}:
    Exclude heavily preprocessed datasets, datasets with inadequate documentation on origin/pre-processing and datasets with an irreversible target leak.
    \item \textbf{No License Issues}:
    Exclude any dataset whose license does not allow for use in an academic benchmark.
    \item \textbf{Open-access Structured Data API}:
    Exclude any datasets which cannot be automatically downloaded from a data repository with structured metadata. 
    \item \textbf{Ethically Unambiguous Tasks}:
    Exclude datasets with tasks that pose ethical concerns
\end{itemize}

\subsection{Models}
As formerly mentioned, our focus is on calibration methods applicable to popular machine learning models. To determine which methods are popular among practitioners, we've synthesized results from surveys addressing usage among practitioners, literature reviews, and studies where machine learning is applied to a problem within a domain. 

Overall practitioners appear to use a mixture of classical-, tree based-, and neural network models, in addition to AutoML. Regression models are sometimes extended to binary problems. Binary models are usually extended to the multi-class setting using "one vs. all" or "one vs. one". Models are also ensembled using bagging, boosting, routing/gating (Mixture of Experts), and stacking \citep{jurek_survey_2014}. 

\begin{itemize}
    \item Classical models
        \begin{itemize} 
        \item Support vector Machines (SVM)
        \item Logistic regression (LR)
        \item Bayesian approaches
            \begin{itemize}
            \item Naive Bayes (NB)
            \item Bayesian Networks (BN)
            \end{itemize}
        \item Linear discriminant analysis (LDA)
        \item Similarity search
            \begin{itemize}
            \item K Nearest Neighbors (K-NN)
            \item Radius Neighbors (R-N)
            \end{itemize}
        \end{itemize}
    \item Tree models 
    \begin{itemize} 
        \item Random Forest (RF)
        \item CatBoost (CB)
        \item XGBoost (XGB)
        \item LightGBM (LGBM)
    \end{itemize}
    \item Artificial Neural Networks 
    \begin{itemize}
        \item Convolutional neural networks (CNN)
        \item Deep neural networks (DNN)
            \begin{itemize}
                \item Transformer networks (Transformers) 
            \end{itemize}
        \item Recurrent neural networks (RNN)
        \item Feedforward neural networks (FNN)
            \begin{itemize}
                \item Multilayer perceptron (MLP) 
                \item Probabilistic neural networks (PNN)
            \end{itemize}
        \item Graph neural networks (GNN)
        \item Topological neural networks (TNN)
        \item Foundation models (FM)
    \end{itemize}
    \item AutoML 
    \begin{itemize}
        \item AutoGluon 
        \item Google cloud AutoML
        \item AWS Sagemaker
    \end{itemize}
\end{itemize}
Note that there are many more methods which do not neatly fall into these categories such as GANs, evolutionary approaches, fitting multivariate random variables, Non-hierarchical clustering and so on. But we'll not address these due to their lacking popularity, which do not necessarily speak of their potential.  

Kaggle's "state of data science and machine learning" surveys provide an initial view of current practice. In 2021 and 2022, the surveys collected approximately 25 000 responses, of which the data from approximately 2000 respondents that report a job title of "data scientist" are used. The 2021 survey shows that the most commonly used methods are: linear or logistic regression, tree-based methods (RF, XGBoost,CatBoost, and LightGBM), CNNs, Bayesian approaches, DNNs, RNNs, transformer networks, GANs and evolutionary approaches (2021). These are often implemented within the frameworks: Scikit-learn, TensorFlow/Keras, pytorch/lightning/fast.ai, or the framework specific to the model as for XGBoost and CatBoost (2022). The 2021 study also showed an increased usage of AutoML solutions such as Google's cloud AutoML offering (2021). \citep{kaggle_kaggle_2021, kaggle_kaggle_2022} \newline

Literature reviews also show similar results. In marketing, Bayesian approaches, neural networks, LDA, tree-based methods, SVM, Non-Hierarhical clustering, ensambles, and K-NN are in use \citep{duarte_machine_2022}. 
In chemistry, we've seen tree-based methods and neural networks such as FNNs, CNNs, RNNs, GNNs and TNNs \citep{shi_machine_2023}. 
In stellar astronomy we see neural networks (ANNs, MLP, CNNs, RNNs, PNNs), tree based methods, SVM, K-NN and Bayesian methods, in addition to fitting multivariate random variables \citep{li_machine_2025}.   
In opinion mining and sentiment analysis we see Naïve Bayes, Bayesian network, Maximum entropy, SVM, Neural network, KNN, decision trees, and rules based models \citep{hemmatian_survey_2019}.
While in ecology we see LDA, a consensus of binomial logistic regression, multinomial regression and SVM,  tree-based methods, and Stagewise Additive Modelling using a Multi-class Exponential (SAMME) loss function \citep{bourel_multiclass_2018}.

Then there's the studies which aim to catalog methods, which highlight decision trees, Bayesian classification, Rule based classification, SVM, Association rule mining and backpropagation (Neural Networks), K-NN, Case Based Reasoning, genetic algorithms, fuzzy logic and Rough Set Approach \citep{paul_comparative_2020}.
Furthermore in  "An overview of machine learning classification techniques" they highlight NB, decision trees, LDA, Logistic regression, K-NN, SVM, and Neural networks \citep{alnuaimi_amer_fah_overview_2024}.

Lastly in "Benchmarking AutoML for regression tasks on small tabular data in materials design" they present SVM, decision trees, BNN (Bagged NN), GBNN (Gradient Boosted NN), WBNN (Wavelet Bagged NN), WGBNN (Wavelet Gradient Boosted NN), and a plethora of AutoML solutions \citep{conrad_benchmarking_2022}.

\subsection{Calibration methods}
To obtain a well-calibrated function \textbf{f}, we've identified two main methods:
\begin{enumerate}
    \item \textbf{Model selection} \newline
    Model selection pertains to the choice of learner, including it's hyperparameters. We consider the choice of loss function as a hyperparameter. Certain models have been shown to generally achieve higher degrees of calibration out of the box. Additionally hyperparameter optimization has been shown to improve calibration. \citep{bostrom_calibrating_2008}  
    \item \textbf{Post-processing} \newline
    Post-processing pertains to any manipulation of the learners output. Such as the normalization function, which maps a learners output onto the probability simplex, or post-hoc calibration which maps from simplex to simplex. \citep{silva_filho_classifier_2023}  
\end{enumerate}

However, recall that the global optimum is best approximated when the entire architecture is optimized as one. Therefore we won't disregard the potential effects of pre-processing techniques or other components, and highlight that under- and over-sampling techniques have been shown to hurt calibration \citep{carriero_harms_2025}. Additionally, combining calibration methods such as hyperparameter tuning and post-hoc calibration techniques have been shown to create solutions better then each technique in isolation \citep{bostrom_calibrating_2008}. Initially, we'll only consider post-hoc calibration methods. 

\subsubsection{Post-hoc calibration}
Post-hoc calibration is the procedure of estimating a function $p$ where $h$ is a learner such that:
\begin{gather*}
    p:\Delta^{k} \rightarrow \Delta^{k} \\
    \textbf{f} = p \ \circ \ h 
\end{gather*}

We restrict attention to model-agnostic post-hoc calibration methods while acknowledging that many model-specific approaches also exist. For empirical evaluation we selected five calibrators from the catalog in \citep{manokhin_machine_2022}: Isotonic regression, Platt scaling, Beta calibration, Venn-Abers predictors, and Pearsonify (a recently proposed method). 

\subparagraph*{Other methods} \label{sec:other-calibrators}
The literature contains a very large and heterogeneous set of alternatives; the list below samples representative families to illustrate the breadth of choices (see \citep{manokhin_machine_2022} for a more comprehensive catalog):
\begin{itemize}
\item \textbf{Binning and histogram methods:} empirical binning, binning averaging, and related piecewise constant estimators.
\item \textbf{Bayesian / GP approaches:} Bayesian calibration maps and Gaussian-process models of the calibration function.
\item \textbf{Tree and ensemble methods:} probability calibration trees, modified logistic model trees, nested dichotomies, and ensemble combinations of scaling and binning.
\item \textbf{Loss / regularization based:} entropy penalties, label smoothing, focal loss, and other objective modifications that alter probabilistic outputs.
\item \textbf{Parametric function fits:} Platt-style sigmoids, exponential fits, polynomials (including constrained/monotonic variants), and other parametric transforms.
\item \textbf{Venn / interval predictors and conformal variants:} Venn-Abers and related conformal/Venn constructions.
\item \textbf{Hybrid and combined strategies:} mixtures or pipelines that apply scaling, binning, and ensembling in sequence.
\item \textbf{Multiclass-specific methods:} temperature (softmax) scaling, vector/matrix scaling, Dirichlet calibration, etc.
\end{itemize}

This diversity means practitioners face many valid, but different, calibration choices; our study focuses on a small popular set. ~\\

\paragraph*{Platt scaling / Platt's method / Logistic regression}~\\
Platt scaling fits a sigmoid (logistic) transform to a classifier's raw scores, producing calibrated probabilities of the form
\[
\hat{p}=\frac{1}{1+\exp(A\cdot s + B)},
\]
where \(s\) is the uncalibrated score and \(A,B\) are fit on a held-out calibration set (usually by maximum likelihood). Implementations are widely available (e.g., scikit-learn). \citep{scikit_learn_logistic}

\paragraph*{Isotonic regression / ROC convex hull method}~\\
Isotonic regression is a nonparametric, monotone mapping from scores to probabilities estimated by pool-adjacent-violators (PAV). It makes minimal functional assumptions (only monotonicity). Variants and related techniques (e.g., ROC convex-hull based adjustments) use piecewise-constant monotone fits to enforce consistency with empirical ROC orderings. \citep{scikit_learn_isotonic}

\paragraph*{Beta calibration}~\\
Beta calibration is a post-hoc, parametric method for binary classifiers that generalizes Platt scaling by fitting a Beta distribution to map scores to probabilities \citep{kull2017beta}. The implementation provides three variants for parameter estimation; in our experiments we used the \texttt{abm} setting, which fits all three parameters of the Beta model.

\paragraph*{Venn-abers predictors} ~\\
Venn‑Abers predictors are conformal‑prediction‑based post‑hoc calibration methods for classification that produces *valid probabilistic estimates* by leveraging the validity guarantees of Venn predictors. Given a scoring classifier and a held‑out calibration set, Venn‑Abers fits two isotonic regression models; one assuming the test instance belongs to class 0 and one assuming it belongs to class 1. Yielding a pair of calibrated probabilities \((p_0,p_1)\) that form a prediction interval for the likelihood of the positive class. This interval reflects uncertainty in the calibrated estimate and is guaranteed to contain the true probability under standard I.I.D.\ assumptions . In practice, this interval is often converted to a single point estimate for decision making using the formula
\[
\hat{p}_{\text{VA}} = \frac{p_1}{1 - p_0 + p_1},
\]
which balances the lower and upper bounds into a single calibrated probability. Venn‑Abers thus combines distribution‑free calibration guarantees with isotonic regression flexibility, providing an alternative to standard post‑hoc methods like Platt scaling or single isotonic regression. The implementation we adopted follows the method described in \citep{vovk2015largescaleprobabilisticpredictorsguarantees}.

\paragraph*{Pearsonify} ~\\
Pearsonify is a lightweight, model‑agnostic Python package for generating classification intervals around predicted probabilities in binary classification tasks \citep{pearsonify2025}. It combines principles from Pearson residuals, a classical statistical measure of goodness‑of‑fit, with conformal prediction to quantify uncertainty in probability estimates without strong distributional assumptions. Given a base probabilistic classifier and a calibration set, Pearsonify produces calibrated lower and upper interval bounds around predicted class probabilities that reflect uncertainty due to limited data and model error; these intervals can then be converted into single point estimates when required.

\subsection{Performance}
%Idea from: https://arxiv.org/abs/2301.10803
The performance of an architecture is a weighted sum or linear combination of its facets. Facets are dimensions or aspects of performance that together determine how well it serves its intended purpose. 
\[\Phi = a_{1}\phi_{1} + a_{2}\phi_{2} + \cdots + a_{r}\phi_{r} 
\] 
While performance is often reduced to a single facet, such as the value of a loss function, accuracy or ROC-AUC, this simplification, though convenient, can obscure important considerations. For instance, cost is an ever-present constraint; few, if any, applications operate with unlimited resources. Latency is another context-dependent concern, some systems require millisecond-level responsiveness, while others can afford to run for hours.
%https://arxiv.org/abs/2301.10803
Though the relevant facets are highly dependent on context, some are archetypical and recur across applications. These include, among other things, predictive or classification ability, discriminatory ability, calibration, and financial impact. Each facet should be quantitatively assessed, yet it is essential to distinguish between the measure used and the concept it aims to capture. The full performance of an architecture is thus best understood as a multi-dimensional construct, shaped by context and by trade-offs between competing goals.

%https://arxiv.org/pdf/2501.19047
A critical pitfall of reducing performance assessment to a single facet is the emergence of trivial solutions that exploit the narrow definition of success. For instance, a system can trivially achieve zero latency and cost by estimating $f$ as a constant function, completely bypassing computation. Similarly, perfect calibration can be achieved by always predicting the empirical class frequencies (ECE), independent of the input \citep{pavlovic_understanding_2025}. While technically optimal under their respective measures, such solutions are useless in practice.
Even improvements in widely valued metrics like accuracy or ROC-AUC may come at unacceptable costs in other facets, such as significant increases in computational load, or inference time. These considerations highlight the importance of assessing performance holistically, rather than optimizing for a single dimension in isolation. While it is often necessary to prioritize specific facets, doing so without considering others can lead to systems that are optimal in theory but impractical in deployment.

Another important yet frequently overlooked facet is operational complexity, which captures how well a method performs in real-world deployment scenarios from an operational point of view. This includes considerations such as runtime stability, robustness to data distribution shifts, susceptibility to technical debt, and ease of integration into existing systems. A method with high operational ease is one that is fast, stable over time, interpretable, and minimally prone to maintenance issues or performance degradation. Thus, a complete evaluation of performance should account for a method’s fitness for production, not just its theoretical or test-set behavior.

It may also be critical to evaluate \textbf{when} $f$ performs well and therefore measure performance under predefined scenarios, or explain differences in performance between scenarios. One could ask whether the $f$ generalizes to out-of-distribution data? How well does it extrapolate and interpolate? Does it perform in sparse vector space or only in dense regions? Does it require large volumes of data to function reliably? In industry, our desire to estimate an optimal $f$ frequently arises due to the potential value generated by improving the assignment of new unobserved instances to its class. By using $f$ instead of some alternative we stand to potentially reap some value gain. Inherently, the generalization of $f$ or it's out-of-sample performance becomes paramount. More abstractly it's the performance in production we desire an estimated measure of, which ought to reflect itself in the evaluation strategy by for example ensuring that the discrepancy between train and test sets represent the discrepancy between train and live-production instances.      

\subsubsection{Classification ability}
With classification ability or predictive ability, we refer to an architecture's ability to correctly assign a class to an instance. To illustrate, we look towards the binary confusion matrix, which designates an output of $f$ aka. $\hat{y}$ as a TP, FN, FP, or TN depending on the true $y$. \citep{geeksforgeeks_confusion_nodate}

\begin{tabular}{ | m{3cm} | m{5cm}| m{5cm} | } 
  \hline
  $y$ \textbackslash \ $\hat{y}$  & Predicted Positive & Predicted Negative \\ 
  \hline
  Actual Positive & True Positive (TP)& False Negative (FN) \\  
  \hline
  Actual Negative & False Positive (FP) & True Negative (TN) \\ 
  \hline
\end{tabular}

TP + TN is therefore the count of correct predictions, FN the count of negative predictions when the true class is positive and so on. From these counts we derive the metrics: Accuracy, Recall, Precision, and F1 Score which we define below:
\begin{gather*}
    Accuracy = \frac{TP+TN}{TP+TN+FP+FN} \\
    Precision = \frac{TP}{TP+FP} \\
    Recall = \frac{TP}{TP+FN} \\
    F1 \ Score = \frac{2 \cdot Precision*Recall}{Precision+Recall}
\end{gather*}

\begin{comment}
%https://medium.com/@sreenilarajesh/confusion-matrix-for-multiclass-classification-fe7b6901a541
These metrics, as defined, are only applicable to binary problems. To extend them to multi-class scenarios, we can use micro- or macro-averaging across classes. 
To micro-average, we iteratively treat each class as a one vs. all binary problem and calculate the confusion matrix to sum the cells across classes.
\begin{gather*}
    Precision = \frac{\sum^{k}_{i=1} TP_{i}}{\sum^{k}_{i=1} TP_{i}+FP_{i}}
\end{gather*}
To macro-average, we iteratively treat each class as a one vs. all binary problem and calculate the metrics to average the metric across classes.
\begin{gather*}
    Precision = \frac{\sum^{k}_{i=1} Precision_{i}}{k}
\end{gather*}
Note that other measures exists such as PR-AUC.
\end{comment}

To visualize overall predictive performance Murphy curves have been suggested \citep{dimitriadis_evaluating_2023}. 

\subsubsection{Discriminatory ability}
Discriminatory ability relates to "what extent the forecast values distinguish situations with lower or higher true event probabilities". It is therefore the order of classes by score
values which is evaluted. To measure discrimination ROC-AUC has been suggested. \citep{dimitriadis_evaluating_2023}

In the binary case, ROC-AUC is the area under the receiver operating characteristic (ROC) curve, which plots TP against FP for all decision boundaries between 0 and 1. According to Dimitriadis et. al. "Since the ROC curve is invariant under strictly increasing transformations of the forecast values, it diagnoses discrimination ability only, while ignoring issues of calibration." \citep{dimitriadis_evaluating_2023}
%Multiclass extension of ROC-AUC

\subsubsection{Calibration}

Calibration is a facet of performance, with varying definitions and measures. Generally, the concept concerns itself with the degree of which class scores are actual probabilities, in a frequenist statistical sense. A class score of 0.8 is considered calibrated if the instance belongs to the given class 80\% of the time when such a score is given. Intuitively, it is therefore the degree of which the vector $\hat{\textbf{y}} = [p_1, \ p_2, \ \cdots, \ p_k]$ is a probability vector. \citep{pavlovic_understanding_2025}

There are multiple ways of formally defining calibration, which vary by the strictness or strength of the definition. We'll consider three alternative definitions: instance-wise, multi-class and class-wise calibration, and corresponding measurements. \citep{shaker_random_2025}

 Recall that under the I.I.D. assumption, each row $(\textbf{x}_{i}, y_i)$ of $T$ is viewed as an independent sample from a random variable distributed as $\mathcal{D}$. This implies the existence of marginal distributions and the conditional probability of class j given instance $\textbf{x}$, $P(\mathcal{Y}=j  \mid \mathcal{X}=\textbf{x})$. $\textbf{f}$ is then considered instance-wise calibrated when: 
 \begin{gather*}
 p_j = P(\mathcal{Y}=j  \mid \mathcal{X}=\textbf{x}) \\
 \forall \ j \in \mathcal{Y} \\ \forall \ \textbf{x} \in \mathcal{X}
 \end{gather*} 
 
In other words, $\textbf{f}$ is truly instance-wise calibrated when it is the true conditional probability function. To measure instance-wise calibration we can use the mean Brier score, Spiegelhalter Z-statistic and mean Log-loss. Brier score and Log-loss are strictly proper scoring rules \citep{silva_filho_classifier_2023}. These two measures are therefore uniquely minimized by the true probabilities. The Brier score is calculated per row $BS(\hat{\textbf{y}}_{i},\textbf{q}_{i})$, and then averaged across the rows $MBS(T)$. 
\begin{gather*}
    BS(\hat{\textbf{y}}_{i}, 
    \textbf{q}_{i}) =\sum_{j=1}^{k}{{(p_j - q_j)}^2} \\
    MBS(T)=\frac{1}{N}\sum^{N}_{i=1}BS(\hat{\textbf{y}}_{i}, 
    \textbf{q}_{i}) \\
    \textbf{f}(\textbf{x}_i) = \hat{\textbf{y}}_{i} \\
    \textbf{q}_i \in \Delta^{k} \\
    q_j \in \textbf{q}_i
\end{gather*}
When $\mathcal{D}$ is known $\textbf{q}_i$ can be derived as:
\begin{gather*}
    q_i(j) = P(\mathcal{Y}=j  \mid \mathcal{X}=\textbf{x}_i)\\
    \textbf{q}_i = [q_i(1), \ q_i(2), \cdots, q_i(k)]
\end{gather*}
When $\mathcal{D}$ is unknown $\textbf{q}_i$ is constructed based on $y_i$, such that the entry corresponding to class $j$ is 1 if $y_i = j$ and $0$ elsewhere:
\begin{gather*}
    \textbf{q}_i = [0,\ 0, \cdots, \ 1,   \cdots, \ 0]
\end{gather*}
\newline

%https://valeman.medium.com/spiegelhalter-z-score-for-binary-classification-25566e43ae0d
%Probabilistic prediction in patient management and clinical trials D. J. Spiegelhalter
Furthermore, in a binary setting the Brier score can be decomposed into a calibration term and a sharpness/resolution term, where the Spiegelhalter Z-statistic evaluates the calibration component of the Brier Score. Many argue that the Brier score conflates calibration and discrimination, whereas Spiegelhalter’s Z focuses purely on calibration.  \citep{spiegelhalter_probabilistic_1986,manokhin_spiegelhalter_2025}

\begin{center}
\begin{gather*}
    p_i = \hat{\textbf{y}}_{i}[1] \ ,  score \ of \ positive \ class \\
    MBS(T) =\frac{1}{N} \sum_{i=1}^{N}{{(y_{i}-p_i)}^2} \\
    MBS(T) = \frac{1}{N} \sum_{i=1}^{N}{{(y_i - p_{i})(1-2p_{i})}} \ + \  \frac{1}{N}\sum_{i=1}^{N}{{ p_{i}(1-p_{i})}} 
\end{gather*}
\text{The first term is the calibration term; the last term is the sharpness term.}
\end{center}

To derive the Spiegelhalter Z-statistic from this decomposition and the corresponding hypothesis test, we first assume that $Y_i \sim Bernoulli(p_i)$ under the null hypothesis. Therefore, the null hypothesis is that the model is calibrated. Under this assumption, these assertions can be proven true:

\begin{gather*}
     E_0(Y_i) = p_i  \\
     E_0(MBS(T)) =  \frac{1}{N}\sum_{i=1}^{N}{{ p_{i}(1-p_{i})}} \\
     V_0(MBS(T)) =  \frac{1}{N^{2}}\sum_{i=1}^{N}{{ (1-2p_{i})^{2}p_{i}(1-p_{i})}} \\ 
     MBS(T) \sim \mathcal{N}(E_0(MBS(T)), V_0(MBS(T))) 
\end{gather*}
Note that $MBS(T)$ approaches the normal distribution in the limit due to the central limit theorem. Furthermore:
\begin{gather*}
    Z = \frac{MBS(T) - E_0(MBS(T))}{V_0(MBS(T))^{0.5}} = \frac{\sum_{i=1}^{N}{{(y_i - p_{i})(1-2p_{i})}}}{\sqrt{\sum_{i=1}^{N}{{ (1-2p_{i})^{2}p_{i}(1-p_{i})}}}} \\
    Z \sim \mathcal{N}(0,1)
\end{gather*}
Where $Z$ is the Spiegelhalter Z-statistic and measures how many standard deviations the brier score deviates from it's expectation under perfect calibration. Low $|Z|$ indicates high calibration, while high $|Z|$ indicates low calibration. By choosing a level of significance we can also perform a hypothesis test of the null hypothesis.

%https://valeman.medium.com/spiegelhalter-z-score-for-binary-classification-25566e43ae0d	
An alternative measure is the log-loss (LL) (a.k.a. binary cross-entropy or negative log-likelihood):
\begin{gather*}
    p_{y_{i}} \ is \ the \ score \ assigned \ to  \ the \ true \ class \\
    LL(\hat{\textbf{y}}_{i}, 
    y_{i}) = -\log(p_{y_{i}})   \\
    MLL(T)=\frac{1}{N}\sum^{N}_{i=1}LL(\hat{\textbf{y}}_{i}, 
    y_{i})  \\
\end{gather*}
Log loss heavily penalizes under-confidence in the correct class, since $\lim_{x\to 0} -\log(x) = \infty$. In contrast, the Brier score penalizes errors quadratically and does not diverge in the interval $[0,1]$, making it less sensitive to large deviations.

Alternative definitions of calibration do not condition on the instances $\textbf{x}$ but on the class scores $p_j$, this is known as probability-wise calibration. First, we define the space of $\hat{\textbf{y}}$ and $p_j$ as:
\begin{gather*}
\hat{\boldsymbol{\mathcal{Y}}} = \{\ \hat{\textbf{y}} \mid \ \mathbf{f}(\mathbf{x}) = \hat{\textbf{y}}, \ \textbf{x} \in \mathcal{X}  \} \\
{\mathcal{P}}_j = \{\ p_j \mid \ p_j  = \hat{\textbf{y}}[j], \ \hat{\textbf{y}}  \in \hat{\boldsymbol{\mathcal{Y}}}  \}
\end{gather*}
$f$ is then considered to be multi-class calibrated when:
 \begin{gather*}
 p_j = P(\mathcal{Y}=j  \mid \hat{\boldsymbol{\mathcal{Y}}}=\hat{\textbf{y}}) \\
 \forall \ j \in \mathcal{Y} \\ \forall \ \hat{\textbf{y}} \in \hat{\boldsymbol{\mathcal{Y}}}
 \end{gather*}
\newline
While $f$ is considered to be class-wise calibrated when:
 \begin{gather*}
 p_j = P(\mathcal{Y}=j  \mid \mathcal{P}_j=p_j) \\
 \forall \ j \in \mathcal{Y} \\ \forall \ p_j \in \mathcal{P}_j
 \end{gather*}
\newline
 Therefore, multi-class and class-wise calibration both condition on class scores. But multi-class conditions on the entire vector of scores while class-wise conditions on score $j$. In binary problems, multi-class and class-wise calibration can be collapsed into one. 
 
 To measure multi-class and class-wise calibration we use specific variants of the expected calibration error (ECE). Many variants of ECE exists, below we'll provide an abstract definition of ECE which encompass multi-class and class-wise definitions \citep{famiglini_towards_2023}.  
 
 The ECE is calculated per class over a dataset $T$. ECE for a dataset $T$ is then aggregated across the classes. The general procedure of calculating ECE is similar across probability-wise calibration defintions. Given a dataset $T$ with $N$ rows calculate $\textbf{f}(\textbf{x}_{i})=\hat{\textbf{y}}_{i}, \forall \ i \in \{1, 2, \dots, N\}$ where $p_{i,j} \in \hat{\textbf{y}}_{i}$ is the class score of class $j$ for instance $i$. Then group the rows $(\textbf{x}_i, y_i)$ based on $\hat{\textbf{y}}_{i}$ when measuring multi-class calibration or $p_{i,j}$ when measuring class-wise calibration, using a fuzzy match:
 \begin{itemize} 
 \item Fuzzy match \newline
 Place rows with similar $\hat{\textbf{y}}_{i}$ or $p_{i,j}$ in the same group. Often achieved by binning/partitioning the probability simplex $\Delta^k$ or the interval $[0,1]$. Many binning procedures have been suggested in the literature such as M equal width binning, the Monotonic Sweep
 Method with equal mass (MSM), and the pool-adjacent-violators algorithm (CORP). 
 \end{itemize}
We'll call each such group a bin $B$ and the collection of bins for $\mathcal{B}$. The number of rows in a bin is $|B|$, such that: \[ \sum_{B \in \mathbf{B}} |B| = N \] Within each bin we calculate the observed frequency of class label j \[ \Bar{p}_{j,B} = \frac{1}{|B|} \sum_{i \in B} \textbf{1}(y_i=j) \] and the average class score as \[ \Tilde{p}_{j,B} = \frac{1}{|B|} \sum_{i \in B} p_{i,j} \]
where the sum $\sum_{i \in B}$ is understood as "for each row in $B$". 
 The ECE of class j over dataset T is then finally calculated as:
\begin{gather*}
    ECE_j(T) = \sum_{B \in \mathcal{B}}{\frac{|B|}{N}}  |\Bar{p}_{j,B} - \Tilde{p}_{j,B}| 
\end{gather*}
While the ECE of dataset $T$ is then finally calculated as:
\begin{gather*}
    ECE(T) = \sum_{j=1}^{k}ECE_j(T)  
\end{gather*}
\
\newline
%https://www.researchgate.net/publication/374297382_Towards_a_Rigorous_Calibration_Assessment_Framework_Advancements_in_Metrics_Methods_and_Use
$ECE(T)$ as defined above is called static calibration error or frequency based ECE. 
%Speficic formulations of the binary ECE variants we'll use

%https://ebooks.iospress.nl/doi/10.3233/FAIA230327
Many authors critique the ECE for being uninterpretable, inconsistent, and noncomprehensive. ECE, is known to be severely affected by the binning scheme. It has also been shown that the ECE is a biased estimator of the true calibration error and the error is strongly dependent on the choice of the number of bins. To address this, the Estimated Calibration Indices (ECI) was made as a modification of the ECE. 

ECI defines a suite of five indices, namely the local calibration index, $ECI_l$, an over-confidence, $ECI_{over}$, and under-confidence, $ECI_{under}$ calibration index, and two global calibration indices, $ECI_{global}$ and $ECI_{balance}$.  $ECI_l$ allows you to measure calibration error in specific regions of the probability space. $ECI_{over}$ and $ECI_{under}$ measures the global tendency toward over- and under-confidence. While, $ECI_{global}$ measures the overall calibration level in absolute terms and $ECI_{balance}$ in terms of overall tendency toward over- or under-confidence.     

To calculate the binary indices we start with binning the rows of $T$ into a collection of bins by partitioning the interval $[0,1]$ and grouping by the class score of the positive class. Per bin, calculate the observed frequency $\Bar{p}_{1,B}$ and the average class score $\Tilde{p}_{1,B}$ as described above. These pairs of values per bin, now constitute a point in the first quadrant of the euclidean plane $\textbf{P}_{B}=(\Tilde{p}_{1,B},\ \Bar{p}_{1,B} )$. If this point lies close to the main diagonal (as in a reliability diagram) the lower the calibration error. Therefore, the point's distance from the line becomes a measure of miscalibration. This distance is found by projecting the point onto the line and calculating the euclidean norm between these points. 
\begin{gather*}
    \langle \cdot,\cdot\rangle \ Dot \ product\\
    |\cdot|_2 \ Euclidean \ norm\\
    \textbf{b} = \frac{1}{\sqrt{2}}(1,1) \\
    \textbf{P}^*_{B} = \langle \textbf{b}, \textbf{P}_{B}\rangle \textbf{b} \\
    d_{B} = |\textbf{P}^*_{B}-\textbf{P}_{B} |_2
\end{gather*}
The distance $d_{B}$ is then normalized by dividing it with the maximum distance from the main diagonal given the same average class score $d^{max}_{B}$. The point with the maximum distance is $\textbf{P}^{max}_{B}=(\Tilde{p}_{1,B},\ \Bar{p}^{max}_{1,B} )$, where: 
\begin{gather*}
\Bar{p}^{max}_{1,B} =  \begin{cases}
    1, & \text{if } \Tilde{p}_{1,B}  \leq 0.5 \\
0, & \text{else}
\end{cases}   
\end{gather*}
$ECI_l$ for bin $B$ can now be calculated as:
\begin{gather*}
    ECI^B_l = 1- \frac{d_B}{d^{max}_B}
\end{gather*}

The other indices are then calculated based on $ECI_l$. Firstly, we collect all the bins $B$ where the scores are under- and over-confident into separate collections. The overconfident set $\mathcal{B}^{-}$ are the bins where:  
$\Tilde{p}_{1,B} >\ \Bar{p}_{1,B}$ and the underconfident set $\mathcal{B}^{+}$ contains $\Tilde{p}_{1,B} \leq\ \Bar{p}_{1,B}$ such that $\mathcal{B} = \mathcal{B}^{-} \cup \mathcal{B}^{+}$ The remaining indices are then calculated as: 
\begin{gather*}
    ECI_{under} = \frac{\sum_{B\in \mathcal{B}^{+}}\frac{|B|}{N}ECI^B_l}{\sum_{B\in \mathcal{B}^{+}}\frac{|B|}{N}}\\ \\
    ECI_{over} = \frac{\sum_{B\in \mathcal{B}^{-}}\frac{|B|}{N}ECI^B_l}{\sum_{B\in \mathcal{B}^{-}}\frac{|B|}{N}}\\ \\
    ECI_{balance} =ECI_{over} - ECI_{under}\\  \\
    ECI_{global} = \frac{\sum_{B\in \mathcal{B}}\frac{|B|}{N}ECI^B_l}{\sum_{B\in \mathcal{B}}\frac{|B|}{N}}
\end{gather*}
ECI is extended to multiple classes using a "one vs. all" scheme and averaging the results per class.

The ECI has been shown to exhibit a lower bias then ECE, on average. The ECI metric also generally over estimates the true calibration error. ECE and ECI are not proper scoring rules and inconsistent. They can be optimized by always predicting the empirical frequencies, leading to misleading calibration measurements in heavily skewed class problems if the model consistently assigns a high score to the frequent class.  

Note there are several other definitions and measures in the literature on calibration, such as confidence calibration, an alternative definition of calibration, or the Hosmer-Lemeshow test, an alternative evaluation of calibration. \newline

To visualize calibration reliability curves have been suggested \citep{dimitriadis_evaluating_2023}.

\subsubsection{Computational cost}
Computational cost relates to the amount of computational resources each architecture consumes, both during training and inference. It is a proxy for the cost accrued when the model is created and subsequently used, but also independently interesting to assess productivity, "performance complexity" and bottlenecks.

Our measures of computational cost are inspired by how companies such as AWS price cloud products such as EC2 instances. Where you pay an hourly rate dependent on the instance type, which designates the amount of computational resources available to the virtual machine. \citep{amazon_web_services_aws_amazon_2025} 

To measure computational cost, we measure:
\begin{itemize}
    \item Training time \\
    Wall time spent training the entire architecture
    \item Inference time \\
    Wall time spent on inference
    \item CPU usage \\
    CPU time spent during training and inference
    \item RAM usage \\
    Peak RAM usage during training and inference. \\ 
\end{itemize}

We measure peak RAM usage since you'd have to provision for the maximum memory footprint to avoid out-of-memory failures. We also record wall-clock and CPU time as proxy measures of training and inference cost. We do not report disk or network usage, not because our methods never utilize them, but because none of them require them during model execution (some architectures could swap to disk to reduce RAM usage, but we have spare RAM).

Downloading and storing model artifacts does incur network bandwidth and disk‐space overhead, but these costs are typically an order of magnitude lower than memory and compute expenses. That said, specialized use cases, such as on-device hosting in mobile or embedded environments could shift the bottleneck toward storage or bandwidth, and merit further investigation.

With regards to costs one ought to remember that the implementation is highly relevant, inefficient implementations of an architecture may degrade it's performance at no fault of the underlying mathematics. Wall time could be high due to lacking parallelization, RAM usage could be high due to inefficient data structures, CPU usage could be high due to thread contention and so on.

Cost must be quantified in any rigorous benchmark study. It's an ever present constraint directly related to any value generated by modeling tabular problems. Just as optimal financial decision making maximizes net present value, architecture choice ought to follow suit. Increasing cost is only justified by marginal value gains, preferably allocated to minimize opportunity costs. Without it, industry practitioners risk becoming cost centers, and academic benchmarkers cannot credibly claim one architecture’s relative superiority. 

As classical algorithms are judged by how their run time scales with problem size (time complexity), architectures should be judged by how their performance scales with available computational resources ("performance complexity"). Formally, if $\mathbf{R}$ denotes our resource budget, methods whose performance scales as $\theta(\log \mathbf{R})$ will asymptotically outperform those scaling as $\theta(R)$, regardless of any constant-factor speedups. Implementing such an evaluation is often hampered by implementations that lack native support for specifying resource budgets (e.g., "max 5 CPU cores, 20 GB RAM, 20 minutes"), by algorithms whose resource usage grows discretely rather than continuously, and difficulties in comparing productivity across datasets $T$.

Relative out-performance in non-cost measures can simply be due to vastly greater resource consumption or superior programmatic implementation. Benchmark studies should at least report resource usage alongside other measures. For instance, contrasting out-of-the-box K-nearest neighbors with CatBoost is misleading without noting that K-NN typically trains in a fraction of the time, freeing cycles for feature selection or hyperparameter search. Furthermore, if TabPFN outperforms CatBoost, that advantage must be contextualized by TabPFN’s extensive pre-training. By explicitly measuring and reporting resource usage; CPU/GPU time, memory/cache footprint, or even energy consumption, studies can establish fair “cost-performance” tiers that reflect real-world trade-offs and drive the field toward architectures with provably better "performance complexity", rather than ever-greater brute-force computation.

\subsection{Evaluation strategy}
The evaluation/validation strategy defines how $T$ is sampled and split to assess an architecture's performance. Specifically, it determines how training and test sets are sampled from $T$.

The most basic of methods is to train and test the architecture on all of $T$. However, we're seldom interested in the performance of an architecture on in-sample instances, but rather on out-of-sample instances, which emulates production environments where architectures encounter new instances drawn from the underlying data distribution. Furthermore, the tendency of generating an overfit $f$ arise, which further defeats the purpose of generalization.   

Therefore a common procedure is to partition $T$ into a training and test set (holdout validation). $f$ is generated by training an architecture on the training test, any hyperparameter tuning or post-hoc calibration is performed using a calibration set drawn from the training set and performance is assessed on the test set. However, since the split is inherently random as is the performance assessment prone to random variance. The process could therefore be repeated using multiple splits, which give rise to $r\times k$-fold cross validation, $k$ designating the number of folds and $r$ the number of repetitions with different random seeds. This method could be further modified by generating each fold through randomized, stratified sampling, grouping by feature values and so on.

Prior literature introduce many validation, resampling or over-/under-sampling procedures. In "The importance of choosing a proper validation strategy in predictive models. A tutorial with real examples" the authors emphasize the importance of splitting the data between a training and test set (generalizability), different cross-validation strategies (predictability), jackknifing and bootstrapping (stability) and permutation test (significance) when generating and assessing a reliable model which are models that exhibit \citep{lopez_importance_2023}:
\begin{itemize}
    \item Stability: The model does not change significantly after removing several samples.
    \item Significance: The response of the model is not due to random chance or overfitting.
    \item Predictability: The model is able to predict correctly (i.e., with high figures of merit). 
    \item Generalizability: The model can be used to predict new samples.
\end{itemize}

The authors believe that a validation strategy should be chosen such that the model that is stable and allows us to predict new samples with high reliability is chosen. "The best model is not the model that best fits the calibration data, or delivers the best results in cross-validation or, more generally spoken, resampling. " 

The cross-validation strategies presented are: Leave-one-out cross-validation, Venetian blinds cross-validation, Cross-validation across data structure and Random subsets. While the method to split between train and test are the Kennard-Stone algorithm, Random test-set validation, and Validation across dataset structure. Note that: "Random test-set validation is unsuitable for imbalanced targets. This must be done with special care, always keeping the proportion of the classes in both training and testing data sets" which is to stratify the random samples by target value. 

The authors also recommend alternatives like bootstrapping, permutation tests and jackknifing as the strategies to follow apart from cross-validation, when class imbalance is too large and there are few instances. Other alternatives include over-/under- sampling however these techniques have been shown to negatively affect calibration.
 
Recent work critiquing the methodology of prior benchmark studies also recommend using k-fold cross-validation such as \citep{tschalzev_unreflected_2025} where the authors argue that 5-fold cross-validation (5CV) consistently improves performance over holdout validation. But this argumentation directly contradicts with \citep{lopez_importance_2023} which critiques the rapid application of validation methods without properly evaluating whether the utilized method is appropriate. One example is the tendency to use leave-one-out cross-validation since it is the validation strategy which returns the highest performance measures, without considering the data structure and the reliability of the model. Though the authors don't divulge on the reasoning for why this proclaimed practice is inappropriate anymore than stating it's an issue.  

Prior benchmark studies such as \citep{erickson_tabarena_2025} have chosen a dynamic approach where a dataset-specific repetition strategy is used and the strategy is as follows: (I) for datasets with less than 2500 samples, use 10 times repeated 3-fold outer cross-validation; (II) for all other datasets, we use 3 repeats.

\subsection{Performance evaluation}
After you've ran an experiment you will possess a large collection of performance measurements across datasets, folds $\times$ repeats, and architectures. Now we'll define how we'll process this information to perform the analysis we need to illuminate our research 
objectives. 

We aim to identify calibration methods which consistently yield well-calibrated outputs. Seeing as we explore post-hoc calibration methods, we need to assess whether these methods improved or degraded performance. Inherently, the performance evaluation becomes relative.

First, let:
\begin{gather*}
    \mathcal{M} = \{ SVM,LR, \cdots\} \ \  \text{The space of models}\\ 
    \mathcal{C} = \{None, Platt, Isotonic, \cdots \} \ \  \text{The space of calibrators. }\\
    \text{None being no post-hoc calibration}\\
    \mathcal{A} = \mathcal{M} \times \mathcal{C} \ \ \ \text{The space of architectures}\\
    \mathcal{S} = \{jm1, MIC, \cdots \}\ \ \text{The dataset suite} \\
    \mathcal{F} = \{1, 2, \cdots, 5\} \ \ \text{The folds}\times\text{repeats} \\
    a \in \mathcal{A}, \ s \in \mathcal{S}, \ f \in \mathcal{F}  \\
\end{gather*}

We define a discrete random variable $X_{a}^{p}$ per architecture $(a)$ and performance measure $(p)$. 
\begin{gather*}
\Omega = \mathcal{S} \times \mathcal{F} , \ \omega \in \Omega\\
X_{a}^{p} : \Omega \rightarrow \{1, 2, \cdots, |\mathcal{A}| \}
\end{gather*}
where $X_a^p(\omega)$ is the ordinal rank of $a$ in ascending order relative to all architectures in $\mathcal{A}$ when evaluated on $\omega = (s,f)$, computed as:
\begin{gather*}
    X_a^p(s,f) = 1 +\sum_{a'\in\mathcal{A} \setminus \{a\}}
    \textbf{1}( \ v_{a'}^{p}(s,f) \prec_p v_{a}^{p}(s,f)) \\
\end{gather*}
where $v_{a}^{p}(s,f)$ is the value of $p$ for $a$ on $(s,f)$ and the  preference relation $\prec_p$ defined as:
\begin{gather*}
    v_1 \prec_p v_2 = \begin{cases}
        v_1 < v_2,  \ \text{if lower is better}\\
        v_1 > v_2, \ \text{if higher is better}
    \end{cases}
\end{gather*}

The expected rank of $a$ across folds and datasets is given by the law of total expectation \citep{devore_modern_2021}:
\begin{gather*}
    \mathrm{E}[X_a^p] = \mathrm{E}[ 
                            \mathrm{E}[ 
                                \mathrm{E}[ \
                                    X_a^p \ | \ S,\ F \
                                ] \ | \ S \
                            ]
                        ]
\end{gather*}

where $S: \Omega \to \mathcal{S}$ is the dataset random variable and $F: \Omega \to \mathcal{F}$ is the fold.
The expectations are empirically estimated as: 
\begin{gather*}
\mathrm{\hat{E}}[ \ X_a^p \ | \ S=s,\ F=f \ ] = X_a^p(s,f) \\
\mathrm{\hat{E}}[\mathrm{\hat{E}}[\ X_a^p \ | \ S,\ F \ ] \ | \ S=s \ ] = \frac{1}{|\mathcal{F}|} \sum_{f\in\mathcal{F}} X_a^p(s,f) \\ 
\mathrm{\hat{E}}[X_a^p] = \frac{1}{|\mathcal{S}|\cdot|\mathcal{F}|}\sum_{s\in \mathcal{S}}\sum_{f\in\mathcal{F}} X_a^p(s,f)
\end{gather*}

$\mathrm{\hat{E}}[X_a^p]$ is a Monte Carlo estimator of the expected rank under the ranking distribution. This estimator assumes a uniform prior over datasets and folds. If the datasets are not sampled I.I.D. from the population of datasets (is representative of the population), the estimator may be biased. Additionally, increasing the number of cross-validation repeats per dataset reduces estimator variance. We compute $\mathrm{\hat{E}}[X_a^p]$ for each architecture and select performance measures to assess who has the highest expected performance by $p$ and call it the expected performance rank of $a$ across folds and datasets. The figure is understood as the expected ranking of $a$ if you randomly chose a fold and dataset in the suite. 

To isolate the effect of post-hoc calibration methods (PHCMS) we define change measures that measure performance deltas relative to the uncalibrated learner. For learner $m \in \mathcal{M}$, PHCM $c\in \mathcal{C}\setminus\{None\} = PHCMS$  , and $\omega = (s,f)$
\begin{gather*}
    \text{Marginal change}: \\
    \Delta_{marg}^{p}(m,c,\omega) = v_{(m,c)}^p(\omega) - v_{(m,None)}^p(\omega)  \\
    \text{Relative change}: \\
    \Delta_{rel}^{p}(m,c,\omega) = \text{sign}\left(\Delta_{marg}^{p}\right) \cdot |\frac{\Delta_{marg}^{p}(m,c,\omega)}{v_{(m,None)}^p(\omega)}| \cdot 100 
\end{gather*}

We define another discrete random variable $Y_{c}^{\Delta^{p}}$ per PHCM $(c)$ and performance delta measure $(\Delta^{p})$. 
\begin{gather*}
\Theta = \mathcal{M \times \mathcal{S}} \times \mathcal{F}, \ \theta \in \Theta\\
Y_{c}^{\Delta^{p}} : \Theta \rightarrow \{1, 2, \cdots, | \mathcal{C}\setminus\{None\}| \}
\end{gather*}
where $Y_{c}^{\Delta^{p}}(\theta)$ is the ordinal rank of $c$ in ascending order relative to all PHCMS when evaluated on $\theta = (m,s,f)$, computed as:
\begin{gather*}
    Y_{c}^{\Delta^{p}}(m,\omega) = 1 +\sum_{c'\in \mathcal{C} \setminus \{None, c\}}
    \textbf{1}( \ \Delta^{p}(m,c',\omega) \prec_p \Delta^{p}(m,c,\omega)) \\
\end{gather*}

The expected rank of $c$ across models, datasets and folds is given by the law of total expectation:
\begin{gather*}
    \mathrm{E}[Y_{c}^{\Delta^{p}}] = \mathrm{E}[ 
                            \mathrm{E}[ 
                            \mathrm{E}[ 
                                \mathrm{E}[ \
                                    Y_{c}^{\Delta^{p}} \ | \ M, \ S,\ F \
                                ] \ | \ M, \ S \
                            ]| \ M]
                        ]
\end{gather*}
where $M: \Theta \to \mathcal{M}$ is the model random variable.

The expectations are empirically estimated as: 
\begin{gather*}
\mathrm{\hat{E}}[ \ Y_{c}^{\Delta^{p}} \ | \ M=m, \ S=s,\ F=f \ ] = Y_{c}^{\Delta^{p}}(m,\omega) \\
\mathrm{\hat{E}}[ \mathrm{\hat{E}}[ \ Y_{c}^{\Delta^{p}} \ | \ M, \ S,\ F \ ] \ | \ M=m, \ S=s \ ] = \frac{1}{|\mathcal{F}|} \sum_{f\in\mathcal{F}} Y_{c}^{\Delta^{p}}(m,\omega) \\ 
\mathrm{\hat{E}}[ \mathrm{\hat{E}}[ \mathrm{\hat{E}}[ \ Y_{c}^{\Delta^{p}} \ | \ M, \ S,\ F \ ] \ | \ M, \ S \ ]| \ M = m] = \frac{1}{|\mathcal{S}|\cdot|\mathcal{F}|}\sum_{s\in \mathcal{S}}\sum_{f\in\mathcal{F}} Y_{c}^{\Delta^{p}}(m,\omega) \\
\mathrm{\hat{E}}[Y_{c}^{\Delta^{p}}]  = \frac{1}{|\mathcal{M}|\cdot|\mathcal{S}|\cdot|\mathcal{F}|}\sum_{m\in\mathcal{M}}\sum_{s\in \mathcal{S}}\sum_{f\in\mathcal{F}} Y_{c}^{\Delta^{p}}(m,\omega)
\end{gather*}

$\mathrm{\hat{E}}[Y_{c}^{\Delta^{p}}]$ is also a Monte Carlo estimator and the same considerations apply, additionally the estimator may be biased if the set of models is un-representative. The figure is understood as the expected ranking of a PHCM if you randomly chose a model, dataset and fold. 

We also want to know the impact each post-hoc calibration method has and therefore produce the distributions of $\Delta^{p}(m,c,\omega)$ aggregated across folds and datasets per model and post-hoc calibrator. Additionally we aggregate across models.

\section{Experiment}
\label{sec:experiment}
\subsection{Reproduction}
To reproduce or review the specifics of our experiment's implementation go to our \href{https://github.com/valeman/classifier_calibration/tree/release-v1.0}{GitHub repository} and follow the instructions in the README.md file and run the project on a host machine running Linux (RHEL) with Intel Xeon E5-2690 CPUs, 32 cores and 512 GB RAM if you want to imitate our machine. 

\subsection{Description}

To determine which calibration methods consistently perform, we'll measure the out-of-box performance of each model on a diverse dataset suite, then we'll apply each calibration method on each model to measure the change in performance. Afterwards, we'll assess whether any calibration methods consistently out-perform across models and datasets.  

\subsubsection{Models}
We'll restrict ourselves to Binary problems, and include the models;
\begin{itemize}
\item Empirical class distribution / Class prior (AVG)
\item Support vector machine (SVM)
\item Linear Discriminant Analysis (LDA)
\item Naïve bayes (NB)
\item Logistic Regression (LR)
\item K-Nearest Neighbours (KNN)
\item RandomForest (RF)
\item Gradient Boosting Classifier (GBC)
\item Histogram Gradient Boosting (HGB)
\item ExtraTrees (EXT)
\item Explainable Boosting Machine (EBM)
\item Catboost (CB)
\item XGBoost (XGB)
\item LightGBM (LGBM)
\item ModernNCA (NCA)
\item TabTransformer (TTRA)
\item TabICL (TICL)
\item TabPFN v2 (TABPFN)
\item TabM (TABM)
\item Multilayer Perceptron (MLP)  
\item Real Multilayer Perceptron (REMLP)
\end{itemize}
Prioritizing implementations from Scikit-learn, TensforFlow/Keras, pytorch/lightning/fast.ai, or the model specific framework. 
 
\subsubsection{Calibration methods}
The calibration methods considered are the post-hoc calibration methods:
 \begin{itemize}
     \item Isotonic regression (isotonic)
     \item Platt scaling (platt)
     \item Beta calibration (beta)
     \item Venn-abers predictors (venn abers) 
     \item Pearsonify. (pearsonify)
 \end{itemize}

\subsubsection{Datasets}
To evaluate these methods we'll apply them to the dataset suite Tabarena-v0.1, which contains 30 binary classification problems. We minimally pre-processes the datasets by encoding all categoricals to unique integers, missing values are encoded to their own group. Non-categoricals are coerced to real numbers, missing values and invalid parsing are set to 0. There is no model specific pre-processing.

\subsubsection{Evaluation strategy}
We'll use the evaluation strategy; $1\times5$-fold cross validation. Four folds are collectively the training set while the remaining held-out fold is the test set. There are no repetitions. Each architecture is therefore generated and evaluated 5 times per dataset. The folds are generated using randomized stratified sampling without replacement. The strata is the target class. When a calibration set is required by an architecture, it will be sampled from the training folds using randomized stratified sampling without replacement, the strata still being the target class.

\subsubsection{Performance measures}
To assess performance on each run we'll measure:
\begin{itemize}
    \item Classification ability
    \begin{itemize}
        \item Accuracy
    \end{itemize}
    \item Discriminatory ability
    \begin{itemize}
        \item AUC-ROC
    \end{itemize}
    \item Calibration
    \begin{itemize}
        \item Brier score
        \item Log-loss
        \item ECI global (MSM)
   
    \end{itemize}
    \item Computational Cost
    \begin{itemize}
        \item Total CPU time during training and inference
        \item Peak RAM usage during training and inference
    \end{itemize}
\end{itemize}

ECE and ECI bins are generated using the Monotonic Sweep Method with equal mass (MSM). ECE is calculated using the frequency based definition. 
Calibration, discrimination and classification is measured on the held-out test set. Class predictions are derived from the scores using the highest score, which is a Bayes classifier when the learner is instance-wise calibrated \citep{cohen_properties_2004}. %https://www.researchgate.net/publication/2937717_Properties_and_Benefits_of_Calibrated_Classifiers  

\subsection{Evaluation of the experiment}
In light of our research objectives, the experiment design imply certain strength and weakness in the choices we've made with regards to the dataset suite, evaluation strategy, space of architectures evaluated and performance measurements. 

The space of methods or space of architectures evaluated encompass binary, popular machine learning models and select post-hoc calibration techniques. This provides the grounds to assess which binary calibration methods tend to out-perform among models explored. But does not provide the ground to claim state of the art. Since, the set of post-hoc calibration techniques is limited. We're also missing components in architectures known to improve performance, such as hyperparameter optimization and model specific pre-processing.  

The Tabarena-v0.1 suite is an actively maintained, distillation of a broad collection of dataset suites. It being maintained, allows us to easily rerun the experiment in the future as the dataset develops. It's also curated in such a way that makes it an ideal match for our research objective. The datasets are required to be I.I.D. and arise from real random distributions. They're also required to reflect classification and regression problems you likely would have applied supervised-tabular machine learning to solve. The only questionable aspects of the curation is the size limit, ethical exclusions and lacking evaluation of diversity and representativeness. 

Very small and very large datasets are excluded, since the authors argue that these dataset require specialized methods to handle optimally. I'd rather these datasets be included in the suite, such that the architecture's ability to handle large and small datasets be evaluated. The authors could've made no such exclusion or created multiple suites. 

Ethically ambiguous tasks are also excluded, when they could've been flagged or obscured. Ultimately, these datasets are tables with numbers which we use to benchmark the performance of architectures. These interventions into the suite may weaken it's representativeness and also any subsequent benchmarking.  

Lastly, we found no real consideration made to the qualitative and quantitative diversity of the resulting suite or it's representativeness. The suite should represent the space of possible datasets, but whether the suite does so has not been evaluated. Intuitively, representativeness seems likely due to the high number of benchmark studies being distilled, but it is mere conjecture. The evaluation is missing which weakens any generalization of results beyond the suite. Since there is no quantitative or qualitative assessment of the diversity of datasets, there is no cause to claim the explored methods perform in general or on those with the present nature. 

Stratified randomized $1\times5$-fold cross validation allows us to repeat the experiment multiple times, aggregating performance measures across runs to asses expectation. Our results should therefore be less prone to random variance. Stratification also ensures each class is present in each set according to their empirical frequency. However, we would've liked if our folds were generated in such a way as to guarantee additional properties. Such as the test sets containing interpolative and extrapolative out of sample instances. Furthermore, instances in dense and sparse vector space. We also see that the chosen evaluation strategy may not be optimal for small and heavily imbalanced datasets. 

Our battery of performance metrics is diverse and measures multiple facets of performance. The same facets are evaluated in multiple ways to capture methodological differences and to ensure our results are comparable with previous studies. The battery lacks a principled foundation, establishing which performance facets are essential and which measurements are optimal. There may also be facets of performance relevant to your domain or application which we've failed to consider. 

We contend that performance measures should be strictly proper scoring rules (e.g., log-loss, Brier score). A core objective when estimating $\textbf{f}$ is to recover the true conditional probability function. Crucially, the true $\textbf{f}$ uniquely minimizes log-loss and brier score. Thus, lower values on these measurements provide evidence of approaching ground-truth probabilities. A property absent for ECE and ECI which are also prone to aforementioned errors. 

Computational resources are measured using cgroupv2, which provides kernel aggregated resource usage attributable to a group of processes. We're therefore confident our measurements are all encompassing and accurate. We expect some overhead to be included in the RAM and CPU measurements.

\subsection{Scope of inference}
Due to the nature of our experiment, as evaluated above, the conclusions we may justifiably infer from it's results are limited. Specifically, no claim to have found state-of-the-art may be made, we've inherently explored optimum over a space of architectures and datasets, and can only evaluate expectation in performance within this space. We may generalize these performance expectations to other datasets and models, since these sets are diverse. But the missing evaluation of Tabarena-v0.1's diversity weaken any generalization across datasets. We do not know whether the suite is representative for datasets in general. 

We therefore believe this experiment provides the grounds to illuminate which post-hoc calibration methods among those explored tend to perform across binary classification datasets and models, but cannot claim to have found the best methods. The experiment is also missing any statistical tests to ensure the results are statistically significant. 
\newpage
\subsection{Future extensions}
In a future iteration of this study we'd like modify the experiment to include:

\begin{itemize}
    \item \textbf{Evaluate the diversity and representativeness of the dataset suite} \newline
    Include an investigation into how one ought to evaluate the diversity of a dataset suite, and subsequently evaluate the representativeness of Tabarena-v0.1. Should be done to strengthen the generalization of results across datasets.
    
    \item \textbf{Expand the performance assessment} \newline
    We should examine performance facets more comprehensively, including developing principled approaches for selecting appropriate metrics within each facet rather than relying on ad-hoc choices. 
    
    Results should be validated using appropriate statistical methods to establish significance and reliability. Potential approaches include the DeGroot and Fienberg calibration test for rigorous calibration assessment, and Friedman and Nemenyi tests to evaluate performance differences across diverse datasets. 

    Our assessment of computational cost should be more comprehensive. RAM utilization through training and inference should be measured, not only the peak. L1, L2 and L3 cache usage could also be measured along with energy consumption.  
    
    Several technical enhancements would improve measurement accuracy. Process isolation using separate cgroups during model training and inference would reduce measurement overhead and improve resource usage measurement. Dictionary-based storage should be replaced with a n-dimensional tensors for more efficient performance measurement storage and analysis. A Streamlit application could enable user-specific analysis and exploration of results.
    
    Beyond aggregate performance metrics, the evaluation should investigate when methods perform well. This conditional analysis would examine performance under different conditions such as dataset characteristics (size, dimensionality, class balance), data quality issues (missing values, noise, outliers), domain-specific constraints (latency requirements, interpretability needs), resource limitations (computational budgets), and instance characteristics (sparse/dense space, extrapolative/interpolative).
    
    \item \textbf{Refine evaluation strategy} \newline
     Though the chosen evaluation strategy has many strengths, maybe other strategies are more efficient for large, small or imbalanced datasets. The held-out sets should also somehow guarantee the presence of out-of-sample interpolative and extrapolative instances as well as instances in dense and sparse vector space. The Kennard-Stone algorithm could be used to generate these sets. Partly, such that one may evaluate how well each architecture handles such cases.  

    \item \textbf{Define computational budgets} \newline
    Fair comparison of architectures requires establishing explicit computational budgets to ensure performance evaluations occur within comparable resource constraints. These budgets should encompass both operational specifications (such as maximum inference time) and resource limitations (including CPU/GPU time, memory usage, and wall-clock time). 
    
    A practical approach involves defining composite resource budgets that specify maximum computational resource usage over a fixed time period (e.g., maximum 2 CPU cores and 20 GB RAM for 30 minutes of training time). Implementation should utilize process isolation by spawning child processes during both training and inference phases, enabling precise budget monitoring and enforcement. This approach provides the capability to terminate architectures that exceed their allocated resources.
    
    Furthermore, when using a fixed‐budget framework with pre‐trained architectures (e.g., foundation models), two precautions are essential. First, because one off pre‑training confers an advantage over from scratch models, you must specify which development stages count toward the budget and should ideally allow all architectures equivalent pre‑training access. Without this parity, any performance gains may reflect budget differences rather than architectural merits. Though, one could question the degree to which practitioners consider costs they do not incur. 
   
    \item \textbf{Find state of the art} \newline
    The study could be modified to provide the ground to make state of the art claims. To do so all of the above must be done, additionally we must; remove model agnostic pre-processing, use GPUs and expand the space of explored architectures.

    The space of proposed and possible architectures has become massive especially, when you consider all possible combinations or compositions. To truly generate the empirical foundation necessary to support state-of-the-art claims today is therefore a considerably larger undertaking then many authors seem to realize.    
    
    To expand the space of explored architectures we could include more learners and calibration methods, especially model specific methods. When doing so one must ensure that each model is applied optimally such that any performance evaluation reflect it's outmost potential on the given dataset by for example applying relevant pre-processing, hyper-parameter tuning, and ensembling. The explored architectures should also encompass those shown to out-perform previously and those claimed to be state of the art. 
            
     \item \textbf{Include a synthetic dataset suite} \newline
    Should create and benchmark on a synthetic dataset suite. By doing so we can calculate the true conditional probabilities and therefore the true calibration error. We can also guarantee the diversity of the dataset suite, and evaluate how methods perform in a broad range of scenarios. 

    \item \textbf{Extend to multiclass} \newline
    Should expand the study to also include multi-class problems and methods.
    
    \item \textbf{Extend to non I.I.D. data} \newline
    The study could be expanded to include non I.I.D. data such as time series. Because, in real systems, distributions often drift over time. This is especially true in domains regarding complex adaptive systems like economics and biology, where the structure of the underlying system evolves dynamically. In such settings, the I.I.D. assumption frequently breaks down. Considering non I.I.D. scenarios, whether through methods from continual/incremental learning, complete re-training or explicit modeling of distribution drift can greatly improve the relevance of the results.
    
    \item \textbf{Provide theoretical explanations} \newline
    Our study is empirical in nature and provides no theoretical explanation for it's observations. Could include such considerations in the future.
\end{itemize}
\newpage
\section{Sources}
\bibliography{references}

\appendix
\section{Additional Plots}
\label{app:plots}
\includepdf[pages=-, scale=0.9]{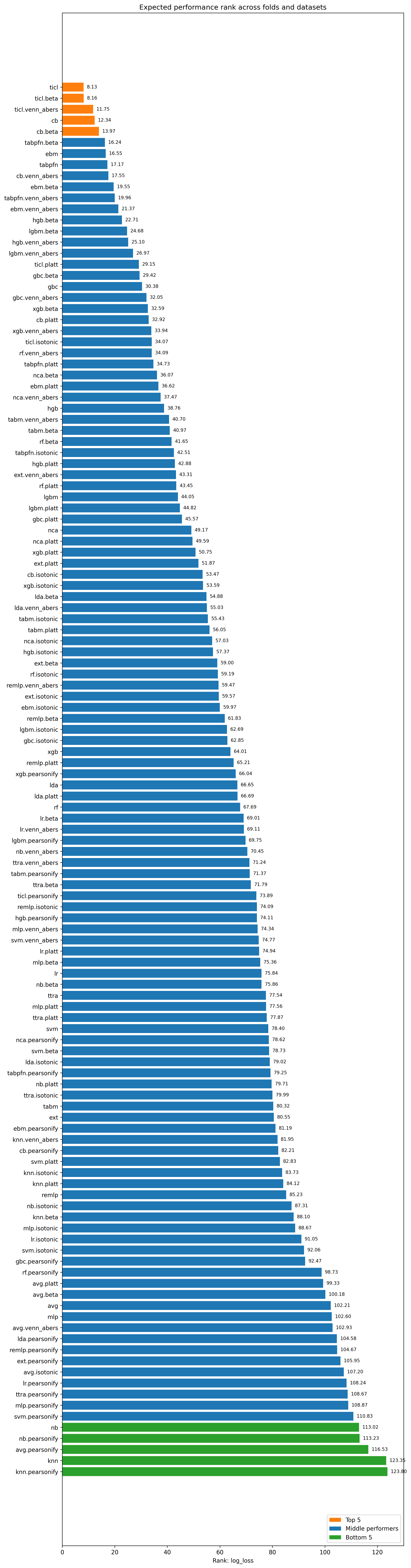}

\end{document}